\documentclass[lettersize,journal]{IEEEtran}
\usepackage{amsmath,amsfonts}
\usepackage{algorithmic}
\usepackage{algorithm}
\usepackage{array}
\usepackage{booktabs} 
\usepackage[caption=false,font=normalsize,labelfont=sf,textfont=sf]{subfig}
\usepackage{textcomp}
\usepackage{stfloats}
\usepackage{url}
\usepackage{verbatim}
\usepackage{graphicx}
\usepackage{multirow}
\usepackage{amssymb}
\usepackage{cite}

\begin{document}

\title{AeroVerse-SatAgent: UAV-Satellite Collaborative Spatial Reasoning Inspired by the Dual Visual Pathway Theory of Cognitive Neuroscience}

\author{Wenyi Zhang, Fanglong Yao,~\IEEEmembership{Member,~IEEE,} Youzhi Liu, Peng Hu, Zhengqiu Zhu, Chen Gao, Xian Sun,~\IEEEmembership{Senior Member,~IEEE,} Kun Fu,~\IEEEmembership{Senior Member,~IEEE}

% \thanks{This work is supported by the National Natural Science Foundation of China under Grants }% <-this % stops a space

\thanks{This work is supported by the National Natural Science Foundation of
China under Grants 62306302, 62425115 and by the Beijing
Natural Science Foundation under Grant L258082.(Corresponding author:
Fanglong Yao.)}
\thanks{Wenyi Zhang, Youzhi Liu, Xian Sun, Kun Fu are with the Aerospace
Information Research Institute, Chinese Academy of Sciences, Beijing
100190, China, and also with the University of Chinese Academy of Sciences, Beijing 100190, China, and with the School of Electronic, Electrical and Communication Engineering, University of Chinese Academy of
Sciences, Beijing 100190, China, and with the Key Laboratory of Target Cognition and Application Technology (TCAT), Aerospace Information
Research Institute, Chinese Academy of Sciences, Beijing 100190, China
(e-mail: zhangwenyi24@mails.ucas.ac.cn; liuyouzhi22@mails.ucas.ac.cn;
sunxian@aircas.ac.cn; kunfuiecas@gmail.com).}
\thanks{Fanglong Yao is with the Aerospace Information Research Institute, Chinese
Academy of Sciences, Beijing 100190, China, and with the Key Laboratory of
Target Cognition and Application Technology (TCAT), Aerospace Information
Research Institute, Chinese Academy of Sciences, Beijing 100190, China (e-mail: yaofanglong17@mails.ucas.ac.cn).}
\thanks{Peng Hu is with the School of Computer Science and Engineering, Beihang University, Beijing 100191, China (e-mail:Misaka\_x86@buaa.edu.cn).}
\thanks{Zhengqiu Zhu is with the National Key Laboratory of Digital Intelligent Modeling and Simulation, National University of Defense Technology, Changsha 410073, Hunan Province, China (e-mail: zhuzhengqiu12@nudt.edu.cn).}
\thanks{Chen Gao is with the Beijing National Research Center for Information Science and Technology (BNRist), Tsinghua University, Beijing, China (e-mail: chgao96@gmail.com).}
}

\markboth{Journal of \LaTeX\ Class Files,~Vol.~14, No.~8, August~2021}%
{Shell \MakeLowercase{\textit{et al.}}: A Sample Article Using IEEEtran.cls for IEEE Journals}


\maketitle

\begin{abstract}
With the rapid advancement of aerospace embodied intelligence, enabling Unmanned Aerial Vehicles (UAVs) to autonomously understand and reason about complex environments has become increasingly important. However, existing UAV-based spatial reasoning approaches face critical limitations: single-view perception renders them vulnerable to occlusions and perspective distortions, while most VLMs lack explicit geometric modeling, relying on semantic cues and yielding inconsistent reasoning under viewpoint and scale variations. To address these challenges, we propose SatAgent, a UAV-Satellite collaborative spatial reasoning model inspired by the dual-pathway mechanism of the human visual system. By jointly leveraging satellite and UAV perspectives, SatAgent enables robust, accurate reasoning in complex urban environments. We first introduce a Geometric-Aware 3D Reconstruction Encoder that elevates 2D UAV features into explicit 3D spatial representations. Next, we design a multi-view topology-semantic alignment module integrating cross-view features within a unified BEV coordinate system. We further introduce a multi-view consistency loss encouraging viewpoint-invariant representations. Finally, we construct SatAgent-SR130K, the first large-scale UAV-Satellite collaborative multi-view spatial reasoning dataset. Experiments show SatAgent outperforms state-of-the-art general-purpose foundation models and specialized spatial reasoning models by 25.91\% and 11.69\%, respectively, across diverse tasks, achieving particularly high accuracy in complex geometric relationship reasoning.
\end{abstract}

\begin{IEEEkeywords}
Multi-view Spatial Reasoning, UAV-Satellite Collaboration, Aerospace Embodied Intelligence, Dual Visual Pathway Mechanism 
\end{IEEEkeywords}

\begin{figure}[!t]
\centering
\includegraphics[width=1\linewidth]{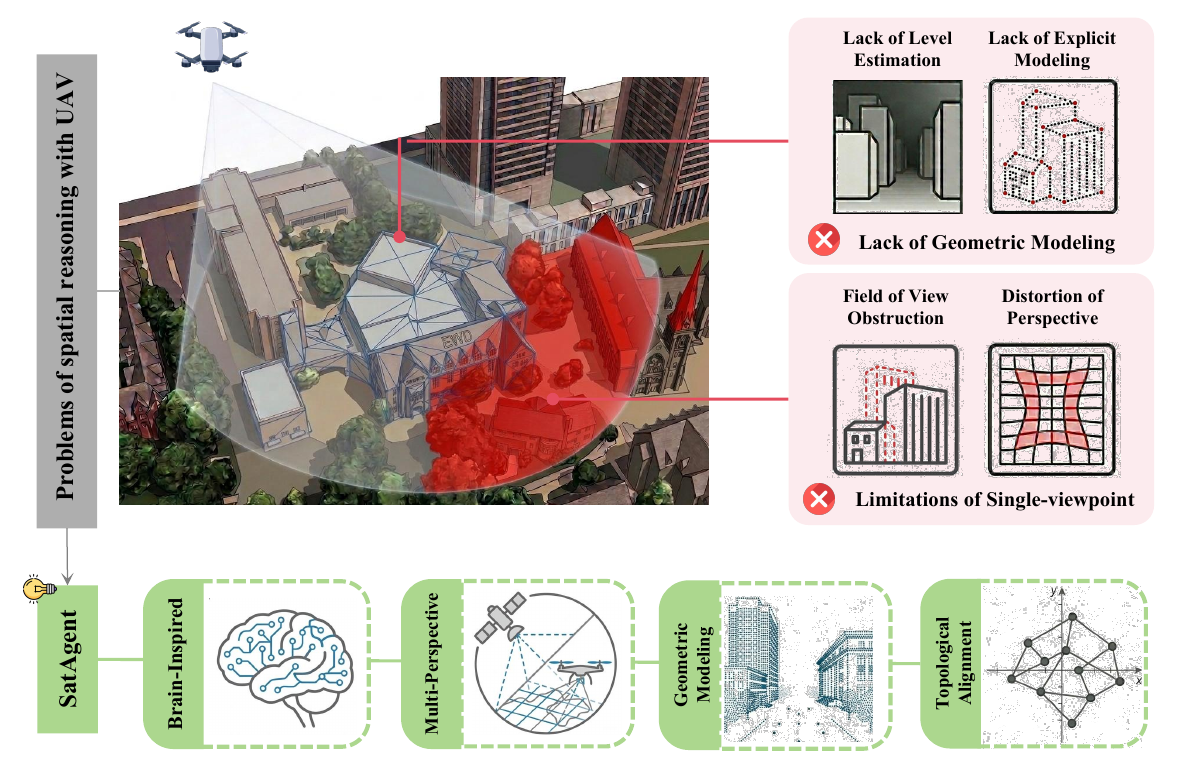}
\caption{Two principal limitations of mainstream UAV spatial reasoning: limitations of single-perspective perception and insufficient spatial geometric structure modeling.}
\label{fig:problem}
\end{figure}

\section{Introduction}
\IEEEPARstart{A}{erospace} embodied intelligence refers to the embedding of perception, cognition, and decision-making capabilities into physical platforms such as satellites and UAVs, enabling autonomous understanding of and intelligent interaction with complex spatial environments \cite{yao2025aeroverse,yao2024aeroverse,liu2024navagent}. Among these, UAVs, as representative low-altitude embodied agents, play an increasingly important role in urban sensing, disaster response, autonomous inspection, and environmental monitoring \cite{johnson2017clevr,anderson2018vision,chang2017matterport3d,mirowski2016learning,kolve2017ai2}. A prerequisite for effectively accomplishing these tasks is that UAVs possess reliable reasoning capabilities over the spatial structures and geometric relationships present in the environment, including the relative positions of buildings, vertical hierarchies, occlusion relationships, and traversable path identification\cite{tan2023knowledge,luo2023depth}. However, real-world scenarios are typically characterized by large spatial scales, complex structures, and drastic viewpoint variations; relying solely on conventional geometric rules or single UAV-perspective inputs is insufficient to meet the spatial understanding demands of complex tasks.
 
In recent years, researchers have explored UAV spatial reasoning from multiple directions, including geometric reconstruction, depth perception, and end-to-end decision-making \cite{kerbl2023gaussian,mildenhall2020nerf,li2024uavnerf,yang2024depthv2,khose2024skyscenes,yue2024semantic,arafat2023navigation,song2023agile}, achieving notable progress. Nevertheless, these methods still fall short of the requirements for high-accuracy spatial reasoning in complex urban scenes, primarily exhibiting the following two limitations, as illustrated in Fig.~\ref{fig:problem}:

\begin{figure*}[!t]
    \centering
    \includegraphics[width=0.85\linewidth]{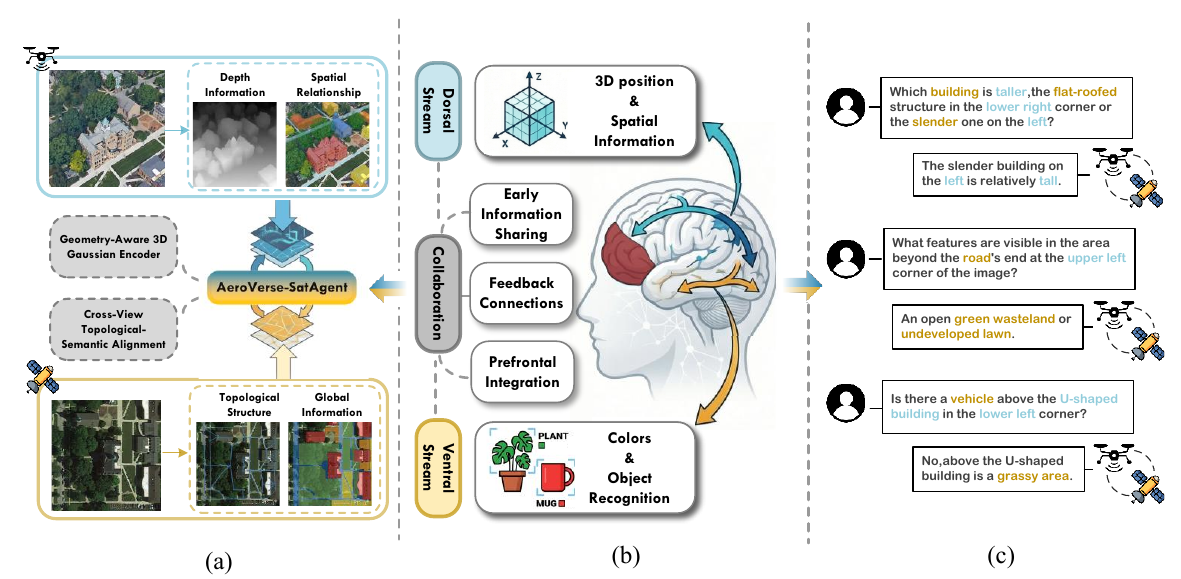}
    \caption{(a) SatAgent jointly leverages the satellite perspective (global
semantic priors) and UAV perspective (local geometric depth) via functionally
complementary dual-branch processing.
(b) The human dual visual pathway: the ventral pathway handles semantic
understanding; the dorsal pathway handles spatial geometry; the two interact
bidirectionally.
(c) Representative SatAgent-SR130K VQA examples requiring UAV-Satellite collaboration for correct spatial inference.}
    \label{fig:intro}
\end{figure*}
 
\IEEEpubidadjcol

\begin{itemize}
    \item \textbf{Limitations of Single-Perspective Perception:} Complex urban environments exhibit pronounced three-dimensional structure and substantial scale variation. Under a single UAV perspective, occlusion, perspective distortion, and depth aliasing are pervasive, forcing the model to simultaneously perform object recognition and spatial relationship reasoning under limited observational conditions, thereby leading to error accumulation. \cite{jia2025omnispatial,liu2023visual,stogiannidis2025mind,cheng2024spatialrgpt}.
\end{itemize}

\begin{itemize}

    \item \textbf{Insufficient Spatial Geometric Structure Modeling:} Existing methods predominantly rely on visual language models for high-level semantic reasoning, but lack explicit modeling of depth, hierarchical structure, and geometric configuration. Spatial relationships are often inferred indirectly through statistical correlations or empirical priors, which tends to fail under viewpoint or scale changes, limiting the geometric consistency and transferability of the reasoning results \cite{shi2022accurate,chen2025spatial,zhang2025sphere,stogiannidis2025mind,wang2024picture}.
\end{itemize}
 
From the perspective of cognitive neuroscience, the remarkable spatial cognitive capabilities of humans do not originate from a single perceptual channel, but rather rely on a multi-pathway cooperative information processing mechanism. The classical dual visual pathway theory \cite{goodale1991neurological,goodale1992separate,ungerleider1982two} posits that the human visual system comprises the ventral pathway and the dorsal pathway: as illustrated in Fig.~\ref{fig:intro}(b), the ventral pathway is primarily responsible for object recognition, categorical discrimination, and semantic comprehension, whereas the dorsal pathway focuses on spatial localization, geometric relationships, depth perception, and action-guidance-related information processing. These two visual pathways form a tightly coupled collaboration through multilevel bidirectional information exchange. The object recognition and semantic features extracted by the ventral pathway serve as target anchors for the spatial localization performed by the dorsal pathway, while the spatial positions and motion trajectories resolved by the dorsal pathway in turn constrain the object segmentation boundaries of the ventral pathway. The two pathways achieve dynamic integration of semantic understanding and spatial reasoning through the interconnections between the parietal and temporal cortices.

Inspired by the human dual visual pathway mechanism, this paper proposes
SatAgent, a UAV-Satellite collaborative multi-view spatial reasoning model.
As illustrated in Fig.~\ref{fig:intro}(a), SatAgent achieves multi-level spatial
understanding by jointly leveraging two complementary perspectives. The satellite
view provides stable top-down global observations suited for characterizing
regional layout, topological structure, and relative positional relationships,
functionally corresponding to the ventral pathway. The UAV view captures rich
local geometry, vertical structure, and depth information through oblique
close-range observations, functionally corresponding to the dorsal pathway.
Mirroring the coordinated ventral--dorsal mechanism, SatAgent maps global
semantic priors from the satellite perspective and 3D geometric cues from the
UAV perspective into a shared coordinate system for joint alignment, yielding
viewpoint-invariant and scale-consistent spatial representations through
structure-aware multi-view relational modeling.

SatAgent introduces three key components: a Dual-Channel Collaborative Encoder,
a Geometric-Aware 3D Reconstruction Encoder, and a Multi-view Topology-Semantic
Alignment Module. Critically, functional separation between the semantic and
geometric branches prevents the two encoders from collapsing into redundant
representations---a common failure mode when heterogeneous visual inputs share
a single feature extractor---while bidirectional cross-stream interaction ensures
that each branch is guided by complementary cues rather than optimizing in
isolation. The reconstruction encoder further reduces the perspective distortion
artifacts inherent in single-view depth reasoning by grounding UAV features in
an explicit metric BEV coordinate system aligned with the satellite view,
improving cross-view spatial correspondence. The alignment module then replaces
naive feature concatenation with structured graph propagation, which captures
non-local topological dependencies that convolution-based fusion consistently
fails to model. Coupled with a multi-view consistency loss providing explicit
gradient supervision across answer generation, structural alignment, and
pathway specialization, the model consistently improves generalization on
tasks requiring metric and topological perception.
 
To support the above research model, this paper constructs SatAgent-SR130K,
the visual question answering dataset specifically designed for
UAV-Satellite spatial reasoning. Built upon GeoText-1652~\cite{chu2024towards}, which
provides geo-referenced scene images with regional descriptions and bounding box
annotations but no question-answer pairs, we introduce 130K spatial reasoning
QA annotations from scratch spanning eight reasoning categories. Beyond scale,
we enforce a single-view unsolvability principle ensuring that
cross-view questions cannot be resolved from either perspective alone, requiring
genuine multi-view collaborative inference rather than single-perspective
shortcuts. We further address a systematic ambiguity absent from the original
dataset: all directional references are unified under the satellite image
coordinate system and annotated via a machine-readable \texttt{direction\_frame}
field, eliminating semantic inconsistencies caused by varying UAV shooting angles.

In summary, the main contributions of this paper are as follows:

(1) We propose SatAgent, a UAV-Satellite collaborative spatial
reasoning model inspired by the dual visual pathway mechanism,
featuring a Dual-Channel Collaborative Encoder that separates semantic
and geometric processing via bidirectional cross-stream gating between
the satellite and UAV branches, establishing a functionally
complementary representational foundation for cross-view spatial
reasoning.

(2) We construct SatAgent-SR130K, the first large-scale UAV-Satellite
collaborative spatial reasoning dataset, comprising 130K QA annotations
across geo-referenced scenes and eight structural reasoning
categories.

(3) We propose a Geometric-Aware 3D Reconstruction Encoder that lifts
UAV features into metric BEV space via covariance-aware 3D Gaussian
soft projection and affine alignment to the satellite coordinate system,
eliminating perspective distortion artifacts inherent in single-view
depth reasoning.

(4) We propose a Multi-view Topology-Semantic Alignment Module
employing dynamic $k$-NN graph attention to model cross-view
topological dependencies, and cross-view gating to adaptively balance
satellite semantic and UAV geometric feature contributions, enabling
structured non-local spatial relationship modeling beyond naive feature
concatenation.

(5) We propose a multi-view consistency loss providing explicit
gradient supervision across three levels: answer generation
($\mathcal{L}_{\text{lm}}$), cross-view structural alignment
($\mathcal{L}_{\text{topo}}$, $\mathcal{L}_{\text{con}}$), and
dual-pathway functional differentiation
($\mathcal{L}_{\text{geo}}$, $\mathcal{L}_{\text{div}}$), ensuring
that the ventral--dorsal functional separation is driven by explicit
training signals rather than implicit emergence.

\begin{table*}[!t]
\caption{Comparison of representative UAV-Satellite Spatial Reasoning Benchmarks.
\textbf{C-V Q.}: dedicated cross-view question split requiring both views
for correct inference;
\textbf{Struct.\ Geo.}: includes dedicated structural geometric reasoning categories (e.g., depth ordering, height--footprint consistency, occlusion completion, path reachability);
\textbf{Ped.}: Pedestrian, \textbf{Veh.}: Vehicle, \textbf{Sat.}: Satellite, \textbf{Tem.}: Templates;
$\checkmark$\,=\,fully supported,
\textit{part.}\,=\,partially supported,
$\times$\,=\,not supported;
$^\dagger$\,approximate value.}
\label{tab:benchmark_survey}
\centering
\renewcommand{\arraystretch}{1.18}
\setlength{\tabcolsep}{3.5pt}
\resizebox{\textwidth}{!}{%
\begin{tabular}{
  >{\centering\arraybackslash}m{3.4cm}
  >{\centering\arraybackslash}m{0.7cm}
  >{\centering\arraybackslash}m{2.6cm}
  >{\centering\arraybackslash}m{1.0cm}
  >{\centering\arraybackslash}m{2.2cm}
  >{\centering\arraybackslash}m{1.1cm}
  >{\centering\arraybackslash}m{1.0cm}
  >{\footnotesize\centering\arraybackslash}m{2.4cm}
  >{\centering\arraybackslash}m{1.0cm}
  >{\centering\arraybackslash}m{1.2cm}
}
\toprule
\textbf{Benchmark}
  & \textbf{Year}
  & \textbf{Platform}
  & \textbf{Input}
  & \textbf{Environment}
  & \textbf{QA Num.}
  & \textbf{Tasks}
  & \textbf{Annotation}
  & \textbf{C-V Q.}
  & \textbf{Struct.\ Geo.} \\
\midrule
\multicolumn{10}{c}{\textit{General Multi-view Benchmarks}} \\[2pt]
VSI-Bench~\cite{yang2024vsibench}
  & 2025 & Egocentric Camera & Video & Indoor
  & ${\sim}5\text{K}$ & 8
  & Human
  & $\times$ & \textit{part.} \\
ViewSpatial-Bench~\cite{li2025viewspatialbench}
  & 2025 & Multi-camera & Image & Indoor + Outdoor
  & ${\sim}5.7\text{K}$ & 5
  & Rules+Tem.
  & \textit{part.} & $\times$ \\
MM-Spatial~\cite{daxberger2025mmspatial}
  & 2025 & RGB-D Camera & Image & Indoor
  & -- & --
  & Rules+Tem.
  & $\times$ & \textit{part.} \\
MMSI-Bench~\cite{yang2025mmsibench}
  & 2025 & Multi-camera & Image & Indoor + Outdoor
  & 1K & 11
  & Human
  & $\checkmark$ & \textit{part.} \\
UrBench~\cite{zhu2025urbench}
  & 2025 & Street+Sat. & Image & Urban Outdoor
  & 11.6K & 14
  & Rules+LLM+Human
  & \textit{part.} & $\times$ \\
All-Angles Bench~\cite{yeh2025allangles}
  & 2025 & Ego+Exo Camera & Image & Indoor + Outdoor
  & 2.1K & 6
  & Human
  & $\checkmark$ & $\times$ \\
\midrule
\multicolumn{10}{c}{\textit{Aerial Multi-view Benchmarks}} \\[2pt]
AirCopBench~\cite{zha2026aircopbench}
  & 2026 & Multi-UAV & Image & Urban Aerial
  & 14.6K & 14
  & \mbox{Rules+LLM+Human}
  & \textit{part.} & $\times$ \\
CityCube~\cite{xu2026citycube}
  & 2026 & Ped.+Veh.+UAV+Sat. & Image &  Aerial + Street
  & 5K & 59
  & Tem.+LLM+Human 
  & \textit{part.} & \textit{part.} \\
Open3D-VQA~\cite{zhang2025open3dvqa}
  & 2025 & UAV & Image & Outdoor Aerial
  & 73K & 7
  & Rules+Tem.
  & $\times$ & \textit{part.} \\
UAVReason~\cite{li2026uavreason}
  & 2026 & UAV & Image & Outdoor Aerial
  & 273K & 22
  & Rules+LLM
  & $\times$ & \textit{part.} \\
LinkS$^2$Bench~\cite{liu2026links2bench}
  & 2026 & UAV+Sat. & Video & Outdoor Aerial
  & 17.9K & 12
  & LLM+Human
  & $\checkmark$ & \textit{part.} \\
\textbf{SatAgent-SR130K (Ours)}
  & 2026 & UAV+Sat. & \textbf{Image} & \textbf{Urban Aerial}
  & \textbf{130K} & \textbf{8}
  & \mbox{\textbf{Tem.+LLM+Human}}
  & $\checkmark$ & $\checkmark$ \\
\bottomrule
\end{tabular}}
\end{table*}

\section{Related Work}
 
\subsection{Spatial Reasoning via Large Language Models}
 
In recent years, multimodal large language models (MLLMs) have achieved
remarkable progress on 2D visual tasks yet continue to face significant
challenges in complex 3D spatial reasoning, including metric distance estimation
and 3D bounding box localization
\cite{cheng2024spatialrgpt,li2025spatialladder,islam2025spatial,tang2025sparkle,liao2024reasoning}.
Early work sought to reconstruct 3D scenes from multi-view images or point
clouds: 3D-LLM \cite{hong2023threedllm} and LLaVA-3D \cite{zhu2024llava3d}
inject the physical world into language models via 3D point cloud features,
but at the cost of substantial computational resources that limit scalability.
ConceptGraphs \cite{gu2024conceptgraphs} reduces this overhead by constructing
3D scene graphs rather than processing raw 3D data directly. A parallel line
develops region-level VLMs to handle complex local spatial relationships:
KOSMOS-2 \cite{peng2024kosmos2}, Shikra \cite{chen2023shikra}, and Ferret
\cite{you2024ferret} achieve region-level understanding via bounding boxes or
masks, while RegionGPT \cite{guo2024regiongpt} and Osprey \cite{yuan2024osprey}
extend this to pixel-level features supporting reasoning over arbitrarily shaped
regions. More recently, approaches enhancing 2D models through monocular depth
estimation have gained traction: SpatialVLM \cite{chen2024spatialvlm} directly
comprehends spatial relationships and metric distances; SpatialRGPT
\cite{cheng2024spatialrgpt} integrates monocular depth into existing visual
encoders, substantially improving directional and distance perception, yet its
spatial representations are still fed into the LLM in textual or symbolic form,
compromising accuracy under complex geometric configurations.

These explorations reveal a fundamental tension: a single perspective cannot
simultaneously achieve precise perception of both global topology and local
geometry. \textit{Motivated by this insight, SatAgent exploits the complementary
advantages of satellite and UAV perspectives, constructing viewpoint-invariant
spatial representations through explicit geometric modeling within a unified
coordinate system, endowing LLMs with metric perception capabilities grounded
in the physical world.}
 
\subsection{Multi-view Spatial Reasoning}
 
Integrating collaborative spatial reasoning from multiple views has emerged
as a key pathway to overcoming the limitation of single-perspective approaches.
Feature-alignment methods such as UCDNet~\cite{Tian2024UCDnet} and
DHD~\cite{Wang2024DronesHelp} achieve UAV-Satellite correspondence via
unsupervised domain adaptation and knowledge distillation, respectively,
but their outputs remain confined to classification labels without spatial
reasoning expressed in natural language.
Graph-based methods such as ConceptGraphs~\cite{gu2024conceptgraphs}
construct unified 3D semantic graphs from multi-view fusion, yet lack
end-to-end learning for viewpoint-invariant spatial relationships.
Multi-image VLMs---MANTIS~\cite{jiang2024mantis} and
LLaVA-NeXT-Interleave~\cite{li2024llava_interleave}---process multiple
visual inputs jointly but rely on implicit spatial representations.
Video-3D LLM~\cite{zheng2025video3dllm} treats video as a multi-view
sequence but degrades under drastic viewpoint changes. BEVFormer~\cite{li2024bevformer} learns unified BEV representations
from multi-camera inputs via spatiotemporal grid queries, yet without
language-grounded spatial reasoning capability.
MM-Spatial~\cite{daxberger2025mmspatial} demonstrates that explicit depth
augmentation substantially improves 3D spatial perception, yet its design
targets indoor small-scale scenes and cannot accommodate the extreme scale
disparities between satellite orthographic and UAV oblique imagery.\textit{In response to these limitations, this paper designs a
Geometric-Aware 3D Reconstruction Encoder and a Multi-view
Topology-Semantic Alignment Module, achieving truly geometrically consistent
multi-view collaborative reasoning.}

Despite these advances, dedicated benchmarks for cross-view aerial spatial
reasoning remain scarce, as summarized in Table~\ref{tab:benchmark_survey}.
General multi-view benchmarks---VSI-Bench~\cite{yang2024vsibench},
ViewSpatial-Bench~\cite{li2025viewspatialbench},
MM-Spatial~\cite{daxberger2025mmspatial}, and
MMSI-Bench~\cite{yang2025mmsibench}---are confined to indoor or mixed-scene
settings and lack aerial-specific geometric reasoning tasks.
Among aerial benchmarks, AirCopBench~\cite{zha2026aircopbench} advances
multi-UAV collaborative perception without global satellite references.
CityCube~\cite{xu2026citycube} spans multiple platforms including satellites
but does not enforce single-view unsolvability, allowing single-perspective
shortcuts.
Most closely related, LinkS$^2$Bench~\cite{liu2026links2bench} addresses
dynamic UAV-satellite cross-view reasoning but focuses on temporal anchoring
via video rather than static structural spatial categories.
\textit{SatAgent-SR130K fills this gap with 130K QA pairs across eight
structural reasoning categories under strict single-view unsolvability
constraints.}

\subsection{Aerial and Remote Sensing Visual Understanding}
 
Research in aerial and remote sensing visual understanding has broadly advanced along two lines. Early work focused on constructing visual question answering benchmarks for remote sensing imagery: RSVQA \cite{Lobry2020} pioneered the introduction of the VQA paradigm into the remote sensing domain, covering basic question types such as existence, quantity, and comparison; however, constrained by template-based generation, it exhibits insufficient semantic diversity, with reasoning depth limited primarily to object-level attribute judgment rather than scene-level spatial relationships. DIOR-RSVG \cite{Zhan2023rsvg} and EarthVQA \cite{Wang2024earthvqa} further incorporate region-level referring expressions and multi-category spatial relationships, providing important benchmarks for fine-grained remote sensing visual understanding evaluation.STAR~\cite{li2024star} provides the first scene graph generation
benchmark over large-size satellite imagery, yet covers only
single-perspective spatial relations. The second line is dedicated to transferring large-scale pre-trained visual language models to remote sensing scenarios: RemoteCLIP \cite{Liu2024remoteclip} substantially improves zero-shot retrieval and classification on remote sensing imagery through continual pre-training on large-scale remote sensing image-text pairs; GeoChat \cite{Kuckreja2024geochat} introduces remote sensing-specific instruction fine-tuning datasets, greatly improving the performance of general MLLMs on scene understanding, object detection, and change detection tasks; SkyEyeGPT \cite{Zhan2024skyeyegpt} incorporates multiple remote sensing visual language tasks into a single model through a unified instruction format, preliminarily exploring cross-task transfer capabilities. However, all of the above works lack systematic exploitation of multi-view collaborative information. To address this gap, University-1652 \cite{Zheng2020university} and GeoText-1652 \cite{chu2024towards} have established geo-referenced image datasets from satellite, UAV, and ground-level tri-perspectives, providing an important data foundation for cross-view geo-localization and natural language-guided UAV navigation. \textit{Building upon these foundations, this paper constructs SatAgent-SR130K, and proposes the corresponding UAV-Satellite reasoning model SatAgent, filling a critical gap in the progression from single-perspective perception to multi-view collaborative reasoning in remote sensing visual understanding.}

\begin{figure*}
    \centering
    \includegraphics[width=1\linewidth]{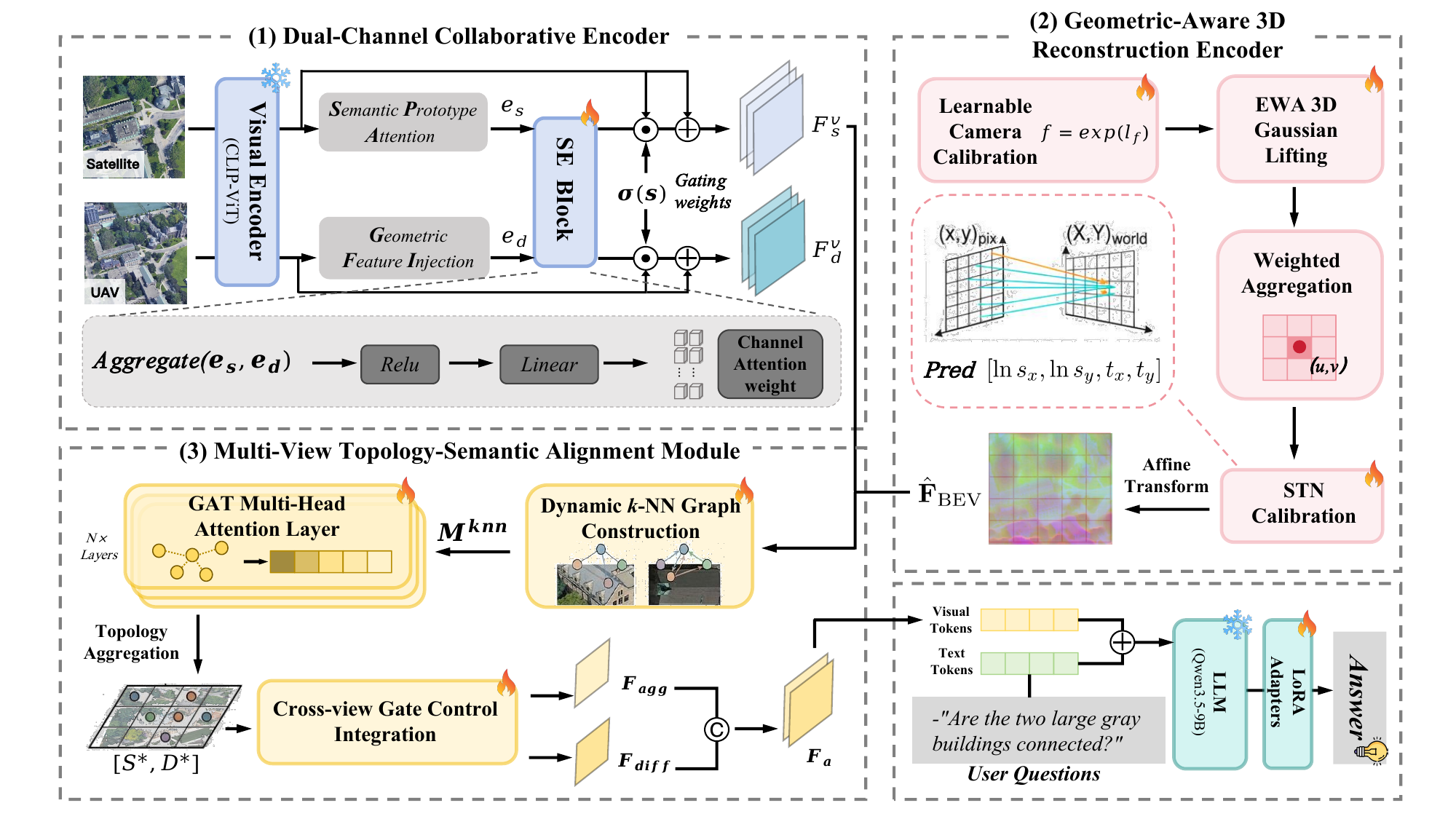}
    \caption{\textbf{Overall architecture of the SatAgent.} The model consists of three core modules working in concert:
    (1) \textbf{Dual-Channel Collaborative Encoder}: Inspired by the human dual visual pathways, it models complementary spatial information through a semantic prototype pathway and a geometric feature injection pathway, and implements bidirectional information feedback via bidirectional cross-stream gating;
    (2) \textbf{Geometric-Aware 3D Reconstruction Encoder}: Utilizes learnable camera calibration to guide EWA 3D Gaussian Lifting, soft-projecting UAV-view features into BEV space, and achieves coordinate alignment at the physical scale via STN affine transformation;
    (3) \textbf{Multi-view Topology-Semantic Alignment Module}: Captures cross-view topological dependencies based on feature semantic similarity via a dynamic k-NN graph and a graph attention network (GAT), and adaptively balances satellite semantic and UAV geometric features via cross-view gating.
    Finally, the aligned visual tokens and the question are jointly input into a LLM for geometry-aware enhanced spatial reasoning and answer generation.}
    \label{fig:pipeline}
\end{figure*}

\section{Method}
 
This section systematically describes the methodological details of the SatAgent model.
\S\ref{sec:arch} presents the overall architecture and data flow;
\S\ref{sec:dual} introduces the Dual-Channel Collaborative Encoder;
\S\ref{sec:gaussian} describes the Geometric-Aware 3D Reconstruction Encoder;
\S\ref{sec:topo} elaborates the Multi-View Topology-Semantic Alignment Module;
\S\ref{sec:llm} presents the LLM inference and parameter-efficient fine-tuning;
\S\ref{sec:loss} details the multi-view consistency loss function design.
 
% ------------------------------------------------------------------
\subsection{SatAgent Architecture}
\label{sec:arch}
% ------------------------------------------------------------------
 
Inspired by the Dual Visual Streams of the human visual system,
SatAgent jointly leverages satellite imagery (providing global topological and semantic priors)
and UAV imagery (capturing local three-dimensional geometry and depth information),
achieving UAV-Satellite collaborative spatial reasoning through a functionally complementary dual-branch
feature processing pipeline.
The overall architecture is illustrated in Fig.~\ref{fig:pipeline}.
 
\noindent\textbf{Overall Pipeline.}
The model sequentially executes the following core stages:
(1) \textbf{Dual-Channel Collaborative Encoder}:
Both satellite and UAV images extract initial features through a frozen CLIP ViT-B/16 \cite{Radford2021CLIP} encoder, which are then processed by specialized adapters configured independently for each branch---the satellite branch, enriched with semantic information, is better suited for region category discrimination,
while the UAV branch, enhanced with geometric injection, is more sensitive to depth structure.
The two branches form a synergistic mechanism of semantic anchoring and geometric constraint through bidirectional gating interaction.
(2) \textbf{Geometric-Aware 3D Reconstruction Encoder} (§\ref{sec:gaussian}):
UAV features are back-projected using a learnable focal length, soft-projected via covariance-aware Gaussian lifting,
and affine-calibrated across views, ultimately achieving spatially meaningful correspondence with satellite features within a unified bird's-eye view (BEV) coordinate system.
(3) \textbf{Multi-view Topology-Semantic Alignment Module} (§\ref{sec:topo}):
The aligned dual-branch features are structure-aware fused via dynamic $k$-NN graph convolution,
converted into visual tokens, and fed into the large language model to complete spatial question answering.
During training, a multi-objective loss jointly constrains the model across three levels---language modeling, structural alignment, and pathway functional separation (§\ref{sec:loss}).
(4) \textbf{LLM Inference} (§\ref{sec:llm}): The aligned visual tokens and the question are jointly input into a large language model (LLM) for geometry-aware enhanced spatial reasoning and answer generation.

\begin{figure*}
    \centering
    \includegraphics[width=1\linewidth]{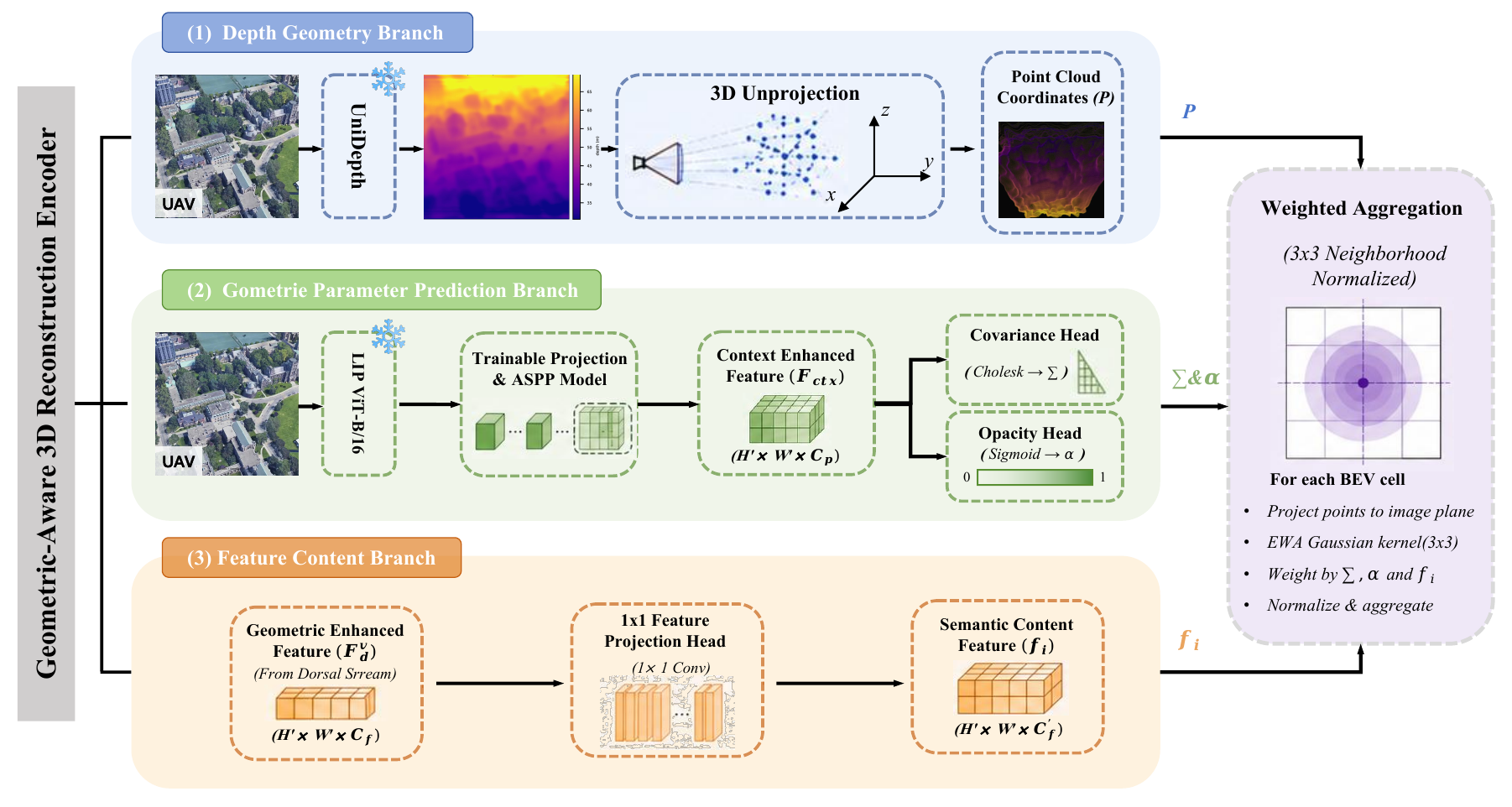}
    \caption{Data flow of the Geometric-Aware 3D Reconstruction Encoder.
Three inputs converge in EWA Gaussian soft projection: the
Depth-Geometry Branch supplies metric 3D point clouds via
frozen UniDepth and a learnable focal calibrator; the Geometric
Parameter Prediction Branch predicts per-pixel Cholesky
covariance and opacity via ASPP; and the Feature Content Input
feeds dorsal stream output $\mathbf{F}_d^v$ as the semantic
payload. Weighted aggregation produces $\mathbf{F}_{\mathrm{BEV}}$,
subsequently aligned to the satellite coordinate system via
STN affine calibration.}
    \label{fig:method1}
\end{figure*}

% ==================== 3.2 Dual-Channel Collaborative Encoder ====================
\subsection{Dual-Channel Collaborative Encoder}
\label{sec:dual}

We design the Dual-Channel Collaborative Encoder to address the
heterogeneity of semantic and geometric features in multi-view
imagery: it extracts semantic prototypes from the satellite view via
a Ventral Pathway Adapter, injects depth-gradient geometry from the
UAV view via a Dorsal Pathway Adapter, and couples the two via
bidirectional gating, establishing a representation that is both
semantically discriminative and geometrically aware.

\subsubsection{Ventral Pathway Adapter}
To make satellite features better discriminate semantic categories
rather than encode local geometry---mirroring the Ventral Stream's
role in object recognition~\cite{goodale1992separate,ungerleider1982two}---we
design a Semantic Prototype Attention (SPA) module that maintains $K$
learnable prototype vectors, simulating category-selective responses
in inferotemporal cortex. The input features activate the
most relevant prototypes via multi-head cross-attention, and the mean
response $\bar{a}_k$ of each prototype aggregates a scene-level
semantic embedding
\begin{equation}
  \mathbf{e}_s = \ell_2\!\left(\sum\nolimits_{k=1}^K \bar{a}_k \mathbf{p}_k\right),
  \label{eq:scene_embed}
\end{equation}
where $\mathbf{p}_k$ is the $k$-th prototype. $\mathbf{e}_s$ then
conditions a channel gate that imposes a semantic directional bias on
the satellite features, while a depth-wise low-pass convolution
suppresses high-frequency geometric noise, reflecting the ventral
pathway's relative insensitivity to fine spatial detail. The output is
the semantically enhanced feature $\mathbf{F}_s^{\delta}$.

\subsubsection{Dorsal Pathway Adapter}
To let the UAV branch perceive depth hierarchy and 3D structure rather
than CLIP's semantic-texture features---corresponding to the Dorsal
Stream's role in depth and spatial relations---we design a geometric
feature injection mechanism. From the existing UniDepth~\cite{piccinelli2024unidepth}
depth map, fixed Sobel/Laplacian operators extract a 6-channel
geometric descriptor combining depth, normalized surface-gradient
directions $(\hat{g}_x,\hat{g}_y)$, gradient magnitude $\hat{m}$,
curvature $\tanh(\nabla^2 D)$, and the interaction term $D\odot\hat{m}$:
\begin{equation}
  \mathbf{G} = \bigl[D,\;\hat{g}_x,\;\hat{g}_y,\;
                     \hat{m},\;\tanh(\nabla^2 D),\;
                     D \odot \hat{m}\bigr].
  \label{eq:geo_feat}
\end{equation}
$\mathbf{G}$ is fused with CLIP features through learnable gating
after lightweight projection, and supplemented with a depth-sensitive
spatial attention conditioned on gradient magnitude, enhancing
responses at object boundaries and depth discontinuities. The output
is the geometrically enhanced feature $\mathbf{F}_d^{\delta}$, together
with a distilled global geometric embedding $\mathbf{e}_d$.

\subsubsection{Bidirectional Cross-Stream Gating}
Inspired by the dorsal--ventral feedback formed through
parietal--temporal interconnections, the two adapters' outputs
($\mathbf{F}_s^{\delta},\mathbf{e}_s$ and $\mathbf{F}_d^{\delta},\mathbf{e}_d$)
are coupled by a bidirectional residual modulation. First, in the
ventral$\to$dorsal direction, $\mathbf{e}_s$ is mapped via SE
attention~\cite{Hu2018SENet} into channel weights for the UAV
features, sharpening geometric responses in semantically salient
regions (e.g., rooftops, intersections); then, in the
dorsal$\to$ventral direction, $\mathbf{e}_d$ correspondingly modulates
the satellite features, using 3D cues to constrain semantic
activations and prevent appearance-only misattribution. Both
modulations are applied residually:
\begin{equation}
  \begin{split}
    \mathbf{F}_d^v &= \mathbf{F}_d^{\delta} \odot
      \bigl(1 + \sigma(s_{v \to d})\cdot \operatorname{SE}(\mathbf{e}_s)\bigr), \\
    \mathbf{F}_s^v &= \mathbf{F}_s^{\delta} \odot
      \bigl(1 + \sigma(s_{d \to v})\cdot \operatorname{SE}(\mathbf{e}_d)\bigr),
  \end{split}
  \label{eq:bidirectional}
\end{equation}
where $\operatorname{SE}(\cdot)$ compresses the embedding to $C/4$
dimensions before restoring $C$ channel weights, and
$s_{v\to d}, s_{d\to v}\in\mathbb{R}$ are gating strengths initialized
to $0$. After gating, $\mathbf{F}_d^v$ flows into the Geometric-Aware
3D Reconstruction Encoder (§\ref{sec:gaussian}), while $\mathbf{F}_s^v$
flows directly into the satellite feature pathway for cross-view
alignment (§\ref{sec:topo}), enabling geometry-aware BEV projection and
semantics-aware alignment, respectively. The residual form keeps
network behavior at initialization equivalent to the ungated case;
training then lets gradients adaptively determine each direction's
activation, yielding functional complementarity rather than forced
symmetry.
 
% ------------------------------------------------------------------

% ------------------------------------------------------------------
 
% ==================== 3.3 Geometric-Aware 3D Reconstruction Encoder ====================
\subsection{Geometric-Aware 3D Reconstruction Encoder}
\label{sec:gaussian}

Inspired by the dorsal stream's role in reconstructing 2D retinal
projections into metric 3D scene representations~\cite{goodale1991neurological,goodale1992separate},
and following the design philosophy of BEVFusion~\cite{liu2023bevfusion},
this encoder combines learnable camera calibration with 3D Gaussian
lifting to map UAV features into a depth-aware BEV representation,
eliminating the perspective gap with satellite imagery (Fig.~\ref{fig:method1}).

\subsubsection{Depth-Geometry Branch and Geometric Parameter Prediction}
Two parallel branches first prepare, for each pixel $i$, the geometric
primitives needed for Gaussian lifting. The \textbf{Depth-Geometry
Branch} uses a frozen UniDepth~\cite{piccinelli2024unidepth} depth map
$\mathbf{D}$ together with a learnable equivalent focal length
$f=\exp(\ell_f)$ and trainable principal-point offsets
$(\delta_x,\delta_y)$ to back-project each pixel $(u,v,d)$ to
\begin{equation}
  X=\frac{(u-c_x+\delta_x)\cdot d}{f},\;
  Y=\frac{(v-c_y+\delta_y)\cdot d}{f},\;
  Z=d,
  \label{eq:unproject_calib}
\end{equation}
yielding $\mathbf{xyz}\in\mathbb{R}^{B\times3\times H_f\times W_f}$,
whose $(X,Y)$ define the BEV location $(u_i,v_i)$ used below. In
parallel, the \textbf{Geometric Parameter Prediction Branch} passes
the CLIP feature map through an ASPP module to obtain
$\tilde{\mathbf{F}}$, from which a covariance head regresses the
Cholesky factor of a positive-definite matrix,
\begin{equation}
  \boldsymbol{\Sigma}_i = \mathbf{L}_i \mathbf{L}_i^\top, \quad
  (L_i)_{jk} =
  \begin{cases}
    \exp(l_{jk}), & j = k \\
    l_{jk},       & j > k,
  \end{cases}
  \label{eq:cholesky}
\end{equation}
while an opacity head outputs $\alpha_i\in(0,1)$. The exponential
diagonal guarantees $\boldsymbol{\Sigma}_i\succ0$ without the
degenerate solutions of direct covariance prediction. Together with
the dorsal-stream feature $\mathbf{F}_d^v$ (the Feature Content Input
in Fig.~\ref{fig:method1}), the quadruple
$(\mathbf{xyz}_i,\boldsymbol{\Sigma}_i,\alpha_i,\mathbf{f}_i)$ fully
specifies the 3D Gaussian element at pixel $i$.

\subsubsection{Covariance-Aware Gaussian Soft Projection}
To let these Gaussian elements genuinely shape the BEV features, we
replace count-normalized hard scatter with an EWA (Elliptical
Weighted Average)-style soft projection~\cite{Zwicker2001EWA}. The
contribution of pixel $i$ to BEV cell $(u,v)$ is
\begin{equation}
  w_{i \to (u,v)} = \alpha_i \cdot
  \exp\!\left(
    -\frac{(u - u_i)^2}{2\sigma^2_{b,x}(i)}
    -\frac{(v - v_i)^2}{2\sigma^2_{b,y}(i)}
  \right),
  \label{eq:ewa_weight}
\end{equation}
where $(u_i,v_i)$ comes from Eq.~\eqref{eq:unproject_calib} and
$\sigma^2_{b,x}(i),\sigma^2_{b,y}(i)$ are the XY components of
$\boldsymbol{\Sigma}_i$ in BEV pixel units. Features are aggregated by
weighted normalization,
$\mathbf{F}_{\text{BEV}}(u,v)=\sum_i w_{i\to(u,v)}\mathbf{f}_i /
(\sum_i w_{i\to(u,v)}+\varepsilon)$, within a $3\times3$ window. This
lets pixels with large depth uncertainty form diffuse responses while
occluded (low-opacity) pixels are automatically suppressed, fully
exploiting the predicted covariance.

\subsubsection{Cross-View Affine Calibration}
Even after focal-length calibration, residual scale and offset
mismatches between the UAV BEV and the satellite coordinate system
persist due to differing ground-sampling distances. We introduce an
STN~\cite{Jaderberg2015STN}-style module that regresses a 4-DoF
transform $\boldsymbol{\Theta}=[\ln s_x,\ln s_y,t_x,t_y]$ from the
globally pooled and concatenated $\mathbf{F}_s^v$ and
$\mathbf{F}_{\text{BEV}}$ via a two-layer MLP (hidden dim $C/4$). The
log-scale parameterization guarantees $s_x,s_y>0$; rotation is
omitted since both views are approximately top-down, and including it
would only add optimization difficulty. The calibrated feature is
obtained via differentiable bilinear sampling,
\begin{equation}
  \hat{\mathbf{F}}_{\text{BEV}}
  = \mathcal{T}\!\left(\mathbf{F}_{\text{BEV}};\,\boldsymbol{\Theta}\right),
  \label{eq:affine_calib}
\end{equation}
with the MLP's final layer zero-initialized for a near-identity start;
$\boldsymbol{\Theta}$ is then optimized jointly by
$\mathcal{L}_{\text{topo}}$ and $\mathcal{L}_{\text{lm}}$, without
extra geometric supervision. Together with the learnable focal length
above, this module constitutes the \textit{BEVScaleCalibrator}
evaluated as Variant~B1 (§\ref{sec:impl}).

\begin{figure}
    \centering
    \includegraphics[width=1\linewidth]{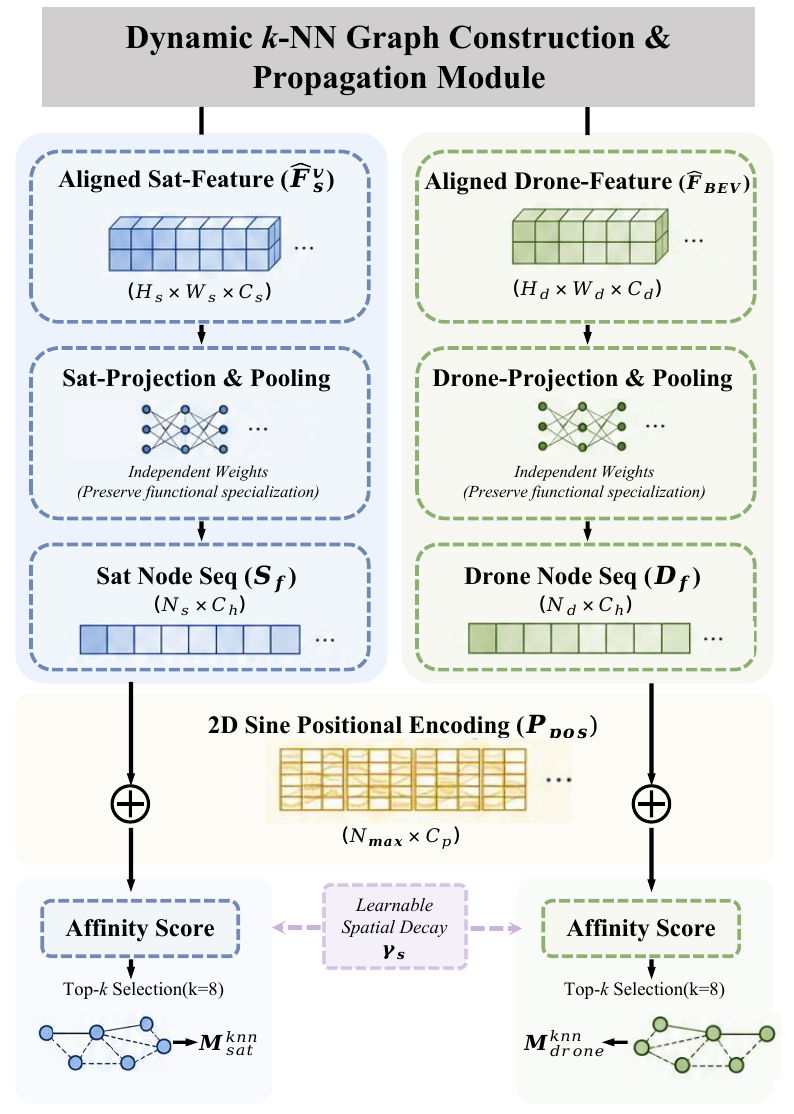}
    \caption{Data flow of the Dynamic k-NN Graph Construction and
Propagation Module. Satellite features $\mathbf{F}_s^v$ and
drone BEV features $\hat{\mathbf{F}}_{\mathrm{BEV}}$ are
independently projected and flattened into node sequences.
Edge affinity scores combine feature cosine similarity with
learnable spatially-weighted positional encoding, from which
the top-$k$ neighbors are retained to form sparse mask
$\mathbf{M}^{knn}$. }
    \label{fig:method2}
\end{figure}
% ------------------------------------------------------------------
% ==================== 3.4 Multi-View Topology-Semantic Alignment Module ====================
\subsection{Multi-View Topology-Semantic Alignment Module}
\label{sec:topo}

When the dual visual pathways converge in prefrontal cortex, a unified
``cognitive map'' associates ventral semantic prototypes with dorsal
spatial coordinates, enabling inference of path connectivity and
regional adjacency~\cite{goodale1992separate,ungerleider1982two}. After
BEV calibration, $\mathbf{F}_s^v$ and $\hat{\mathbf{F}}_{\text{BEV}}$
are geometrically aligned but still semantically divergent---satellite
features carry global topology and regional attributes, UAV features
emphasize local geometry. Since conventional fusion cannot capture
nonlinear cross-view relations such as path connectivity, and
fixed-topology graphs only aggregate spatially adjacent nodes, we adopt
a dynamic $k$-NN graph whose topology is built on-the-fly from feature
similarity, combined with a graph attention network and cross-view
gated fusion (Fig.~\ref{fig:method2}).

\subsubsection{Semantically Adaptive Topology Layer}
To encode the prior that semantically similar regions should be
preferentially connected, the graph is built dynamically from feature
cosine similarity augmented with a learnable spatial-distance bias:
\begin{equation}
  a_{ij} = \frac{\mathbf{h}_i^\top \mathbf{h}_j}{\|\mathbf{h}_i\|\|\mathbf{h}_j\|}
           + \sigma(\gamma_s) \cdot
             \frac{\mathbf{p}_i^\top \mathbf{p}_j}{\|\mathbf{p}_i\|\|\mathbf{p}_j\|},
  \label{eq:knn_sim}
\end{equation}
where $\mathbf{h}_i$ is the node feature, $\mathbf{p}_i$ is a
parameter-free 2D sinusoidal positional encoding~\cite{Vaswani2017Attention},
and $\sigma(\gamma_s)$ is a learnable spatial decay coefficient. For
each node, the $k{=}8$ highest-similarity neighbors are kept (others
set to $-\infty$), giving a sparse mask $\mathbf{M}^{\text{knn}}$, over
which GAT~\cite{Velickovic2018GAT}-style multi-head attention
propagates and updates node features---yielding the
topology-propagated satellite and UAV representations $\mathbf{S}^*$
and $\mathbf{D}^*$ used below. Positional encoding matters because,
after ventral adaptation, two spatially distinct but visually similar
regions (e.g., rooftops of different buildings) could otherwise be
erroneously connected by semantic similarity alone; positional
encoding lets the attention distinguish them.

\subsubsection{Cross-View Gated Fusion}
To adaptively balance satellite semantic information and UAV
geometric information when fusing $\mathbf{S}^*$ and $\mathbf{D}^*$,
we replace simple concatenation---which weighs all channels
uniformly---with channel-wise gating that learns input-dependent
weights emphasizing whichever branch is locally more informative, a
principle validated for heterogeneous multimodal fusion by
Arevalo~et~al.~\cite{arevalo2017gmu}. A differential residual term
further exposes the complementary discrepancy between branches:
\begin{equation}
  \mathbf{F}_a = \operatorname{Conv}_{1\times1}\!\bigl(
    \mathbf{G} \odot \mathbf{S}^* + (1-\mathbf{G}) \odot \mathbf{D}^*\;;\;
    \mathbf{S}^* - \mathbf{D}^*
  \bigr),
  \label{eq:gated_fusion}
\end{equation}
where $\mathbf{G}=\sigma(\mathbf{W}_g[\mathbf{S}^*;\mathbf{D}^*])$ is
the channel-wise gate and $\mathbf{S}^*-\mathbf{D}^*$ supplies a
complementary discrepancy signal. The satellite and UAV branches use
separate, non-shared projection layers, since their feature
distributions have already diverged systematically during the
dual-channel adapter stage---sharing weights here would impose
unnecessary alignment pressure and negate the gains from functional
separation. The resulting $\mathbf{F}_a$ is the aligned feature passed
to the LLM inference stage (§\ref{sec:llm}).

% ------------------------------------------------------------------
\subsection{LLM Inference and Parameter-Efficient Fine-tuning}
\label{sec:llm}
% ------------------------------------------------------------------
 
To effectively inject the aligned visual representations into the large language model while avoiding catastrophic forgetting of the world knowledge carried by the LLM,
we adopt a strategy combining soft token injection with LoRA \cite{Dettmers2023QLoRA} parameter-efficient fine-tuning.
 
The aligned feature $\mathbf{F}_a$ is projected to $N_v{=}256$ visual tokens, and prepended to the question text embedding along the sequence dimension as $[\mathbf{V};\mathbf{T}]$.
The prepended visual tokens enable the LLM to directly access the global visual context during self-attention computation over text tokens.
The LLM backbone (Qwen3.5-9B) is loaded with 4-bit double quantization in NF4 format.
LoRA adapters are applied only to the attention projection layers and feed-forward network layers,
with rank $r{=}16$, making the number of trainable parameters approximately $1\%$ of the total.
During training, labels for question tokens are masked to $-100$,
ensuring that gradients are back-propagated only from the answer generation stage,
preventing the model from memorizing input formats rather than learning visual-text semantic mappings.
Structured output constraints are embedded in the prompts according to question category
(e.g., yes/no questions or single-word/phrase questions),
eliminating the need for constrained decoding on the inference side
and maintaining a streamlined and efficient inference pipeline.

\subsection{Multi-View Consistency Loss Constraints}
\label{sec:loss}

The dual-pathway collaborative mechanism requires explicit training signals
to drive the two-branch features toward functional differentiation.
We jointly constrain learning across three levels---answer generation,
cross-view structural alignment, and dual-pathway functional
differentiation---via the following loss design.

The primary supervision signal is the autoregressive language modeling
loss $\mathcal{L}_{\text{lm}}$, backpropagating gradients only through
answer tokens. At the global embedding level, a symmetric InfoNCE
contrastive loss~\cite{Oord2018CPC,Radford2021CLIP} $\mathcal{L}_{\text{con}}$
(temperature $\tau{=}0.07$) drives viewpoint-invariant scene-level
representations. We additionally propose the following core losses.

\textbf{Position-Aware Structural Consistency Loss.}
Standard Gram matrix matching discards spatial position information and
becomes invalid under the extreme viewpoint disparity between satellite
orthographic and UAV oblique imagery.
Instead, we partition the calibrated BEV feature maps
$\mathbf{F}_s^v,\hat{\mathbf{F}}_{\text{BEV}}$ uniformly into
$R{\times}R$ ($R{=}4$) non-overlapping regions, extract the
$[\mathrm{mean},\mathrm{std}]$ descriptor
$\mathbf{d}_r^v \in \mathbb{R}^{2C}$ per region, and apply symmetric
InfoNCE to the $\ell_2$-normalized descriptor matrices
$\hat{\mathbf{D}}_s,\hat{\mathbf{D}}_d \in \mathbb{R}^{(BR^2)\times 2C}$:
\begin{equation}
  \mathcal{L}_{\text{topo}} = \tfrac{1}{2}\Bigl[
    \operatorname{CE}\bigl(\mathbf{S},\mathbf{y}\bigr)+
    \operatorname{CE}\bigl(\mathbf{S}^\top,\mathbf{y}\bigr)
  \Bigr],
  \label{eq:topo_loss}
\end{equation}
where $\mathbf{S}=\hat{\mathbf{D}}_s\hat{\mathbf{D}}_d^\top/\tau_{\text{topo}}$
($\tau_{\text{topo}}{=}0.1$) and $\mathbf{y}$ assigns each row its
positive-pair identity index. Incorporating region indices as explicit
positional identifiers ensures that positive pairs always correspond to
the same spatial location across perspectives, providing directional
alignment gradients even under large viewpoint disparities.

\textbf{Dual-Pathway Functional Specialization Loss.}
To drive functional separation via explicit gradient pressure rather than
implicit emergence, we design two asymmetric self-supervised losses.

The \emph{Geometric Specialization Loss} $\mathcal{L}_{\text{geo}}$ acts
exclusively on the UAV branch, using UniDepth depth maps as
zero-annotation supervision. A pairwise ranking hinge loss constrains
dorsal-pathway BEV feature magnitudes to be consistent with depth ordering:
\begin{equation}
  \mathcal{L}_{\text{geo}} =
  \frac{1}{|\mathcal{Q}|}
  \sum_{(i,j)\in\mathcal{Q}}
  \max\!\bigl(0,\;
    \Delta - \operatorname{sgn}(D_i{-}D_j)\cdot
    (\|\mathbf{f}_i\|_2 - \|\mathbf{f}_j\|_2)
  \bigr),
  \label{eq:geo_loss}
\end{equation}
where $\mathcal{Q}$ samples pixel pairs with $|D_i{-}D_j|{>}0.05$ and
$\Delta{=}0.1$. The same constraint is deliberately withheld from the
satellite branch, generating asymmetric geometric gradient pressure across
the two branches.

The \emph{Inter-Stream Complementarity Loss} $\mathcal{L}_{\text{div}}$
prevents the two branches from collapsing into redundant representations.
Based on the Barlow Twins~\cite{Zbontar2021Barlow} cross-correlation
matrix $\mathcal{C}$ between batch-normalized embeddings
$\bar{\mathbf{e}}_s$ and $\bar{\mathbf{e}}_d$:
\begin{equation}
  \mathcal{L}_{\text{div}} =
  \sum_c (1-\mathcal{C}_{cc})^2 +
  \lambda_{\text{off}}\!\sum_{c\neq c'}\mathcal{C}_{cc'}^2,
  \label{eq:div_loss}
\end{equation}
where $\lambda_{\text{off}}{=}1.0$. This simultaneously drives same-scene
alignment (diagonal entries $\to 1$) and inter-branch orthogonality
(off-diagonal entries $\to 0$), guaranteeing that the dual-branch design
does not degenerate into single-branch redundancy.

\textbf{Total Loss} combines all signals:
\begin{equation}
  \mathcal{L} =
  \mathcal{L}_{\text{lm}}
  + \lambda_1\mathcal{L}_{\text{topo}}
  + \lambda_2\mathcal{L}_{\text{con}}
  + \lambda_s\bigl(\mathcal{L}_{\text{geo}}+\mathcal{L}_{\text{div}}\bigr),
  \label{eq:total_loss}
\end{equation}
following common multi-task weighting practice~\cite{chen2018gradnorm,kendall2018multitask},
$\lambda_s{<}\lambda_1{<}\lambda_2$ reflects a coarse-to-fine rationale:
global contrastive alignment tolerates larger weight, regional
structural alignment is moderate, and specialization losses act as
light regularizers requiring the smallest weight. So we set $\lambda_1{=}0.1$, $\lambda_2{=}0.2$, $\lambda_s{=}0.05$, with a
500-step warm-up for $\mathcal{L}_{\text{geo}}$ and $\mathcal{L}_{\text{div}}$.

\begin{figure*}
    \centering
    \includegraphics[width=1\linewidth]{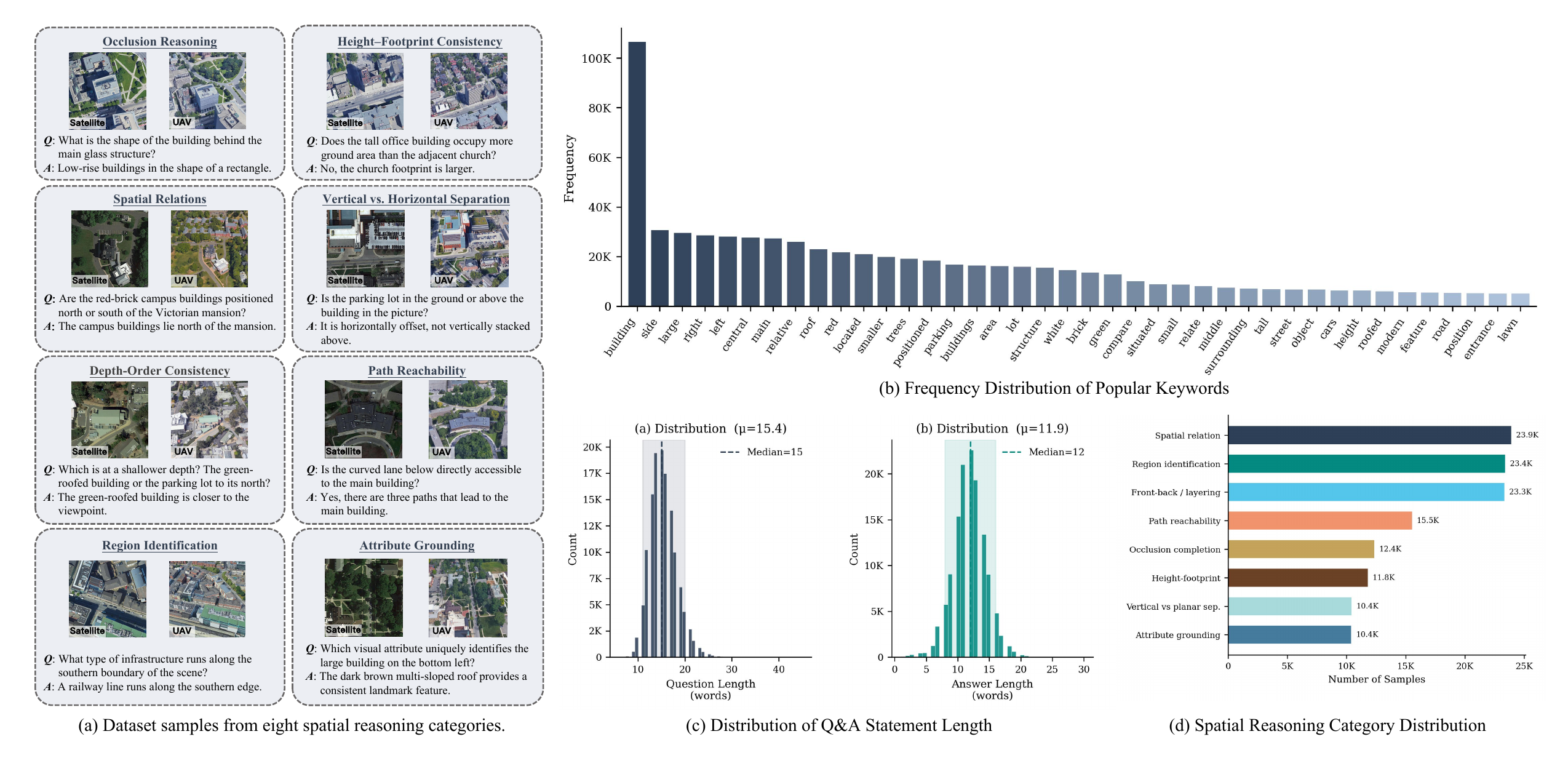}
    \caption{Overview of the SatAgent-SR130K dataset:(a) Dataset samples from eight spatial reasoning categories; (b) vocabulary distribution; (c) length distribution of question-answer sentences; (d) distribution of eight spatial reasoning categories.}
    \label{fig:dataset_intro}
\end{figure*}

\section{Experiments}
\subsection{UAV-Satellite Spatial Reasoning Benchmark}
At present, despite the remarkable progress of visual language models on general scene understanding tasks, dedicated benchmarks for cross-view spatial reasoning remain scarce. Existing visual question answering datasets focus primarily on object recognition and attribute judgment from a single perspective, lacking systematic evaluation of multi-view geometric consistency, depth relationships, and topological connectivity. To address this deficiency, we develop SatAgent-SR130K, the first visual question answering dataset specifically designed for UAV-Satellite multi-view spatial reasoning (Fig.~\ref{fig:dataset_intro}). Built upon GeoText-1652, the dataset encompasses 1,652 geo-referenced scenes. Each scene contains one orthographic satellite image, multiple oblique UAV images captured at varying altitudes and azimuth angles, and ground-level street view images; each image is accompanied by a global description text, a set of region-level descriptive sentences, and bounding box annotations in the normalized format $[\mathit{c\_x}, \mathit{c\_y}, w, h]$, establishing precise correspondences between textual phrases and spatial image regions.
 
To fully exploit all available annotated scenes, we perform a random re-partitioning across all scenes into training, validation, and test sets at different ratios, while ensuring strict geospatial separation between the model development and final evaluation phases. Question-answer pair generation is conducted through an automated pipeline with GPT-5 Nano, employing three complementary generation strategies: (1) \textit{bbox-grounded generation}, which uses annotated region description sentences and their bounding box coordinates as anchors to guide the LLM in generating reasoning questions grounded in specific spatial regions; (2) \textit{caption-level generation}, which jointly inputs the global image description and all region descriptions to generate questions requiring relational reasoning across multiple annotated regions; and (3) \textit{cross-view generation}, which simultaneously provides scene descriptions from both satellite and UAV perspectives to generate questions that can only be answered by integrating both viewpoints, such as identifying architectural structures occluded in the orthographic view but visible in the oblique view. To prevent directional semantic ambiguity arising from varying UAV shooting angles, all directional terms (left/right/up/down) are unified under the satellite image coordinate system (image left corresponds to geographic west, right to east, top to north, and bottom to south); each record is accompanied by a \texttt{direction\_frame} field that annotates this convention in a machine-readable format.
 
To systematically evaluate cross-view spatial understanding capabilities, the designed question-answer pairs span eight predefined spatial reasoning categories: Occlusion Reasoning, Height--Footprint Consistency, Path Reachability, Vertical vs. Horizontal Separation, Depth-Order Consistency, Region Identification, Spatial Relations, and Attribute Grounding. Examples from the eight predefined spatial reasoning categories are illustrated in Fig.~\ref{fig:dataset_intro}. All generated question-answer pairs undergo a three-stage quality filtering process: in the first stage, category labels are verified and automatically corrected for entries that do not conform to the standard set; in the second stage, lexical overlap scores between answer vocabulary and the vocabulary of corresponding global descriptions are computed (excluding stop words), with low-confidence answers subject to manual review to reduce the risk of hallucination; in the third stage, near-duplicate questions within the same scene are detected and only the more elaborately phrased version is retained. Furthermore, all question-answer pairs in the test set and a proportionally sampled subset of the validation set are included in a manual review queue to further ensure the quality of the evaluation data.

\subsection{Evaluation Metrics}
\label{sec:metrics}
We report four complementary metrics spanning lexical overlap to
embedding-based semantics. \textbf{Token F1} measures order-independent
bag-of-words overlap. \textbf{ROUGE-L}~\cite{lin2004rouge}, based on the
longest common subsequence, is order-sensitive. \textbf{METEOR}~\cite{banerjee2005meteor}
adds stemming and WordNet synonym matching with order-aware alignment,
tolerating lexical variation. \textbf{BERTScore-F1}~\cite{zhang2020bertscore}
computes contextual-embedding cosine similarity to capture semantic
equivalence beyond surface form; we apply its \emph{baseline rescaling}
to counteract the anisotropy-induced compression of raw scores into a
narrow high range, expanding the metric's effective dynamic range.

Critically, all four metrics compare against the \emph{reference answer},
not the question. A degenerate response that merely restates the
question or produces fluent but non-responsive meta-commentary shares
little lexical or semantic content with the factual reference, and is
thus penalized across \emph{all four} metrics: near-zero Token
F1/ROUGE-L/METEOR from absent content-word overlap, and low BERTScore-F1
from the embedding distance between question restatement and concrete
answer. This guards against reward hacking by fluency alone, ensuring
measured gains reflect genuine answer correctness.

To assess statistical significance, we additionally compute 95\% confidence intervals via bootstrap resampling (1,000 iterations) over the test set; full results are reported in Appendix~\ref{app:ci}. SatAgent's gains over the strongest baseline (SpatialRGPT, +11.69\% Token F1) exceed the corresponding CI half-widths across all four metrics, confirming the improvements are statistically significant rather than sampling noise.
 
% ============================================================
%  4.1  Implementation Details
%  Dependencies: amsmath, amssymb, bm, booktabs
% ============================================================

\begin{table*}[!t]
\caption{Performance comparison on the SatAgent-SR130K test set.
\textbf{Overall} reports four complementary metrics (Token F1, ROUGE-L,
METEOR, BERTScore-F1); \textbf{By Reasoning Category} reports Token F1.
\textbf{Bold}: best;
\underline{Underline}: second best;
Category abbreviations ---
\textbf{Path}: Path Reachability;
\textbf{Spat.}: Spatial Relations;
\textbf{Attr.}: Attribute Grounding;
\textbf{Reg.}: Region Identification;
\textbf{D.\,-O.}: Depth-Order Consistency;
\textbf{Occ.}: Occlusion Reasoning;
\textbf{H.\,-F.}: Height--Footprint Consistency;
\textbf{V.Sep.}: Vertical vs. Horizontal Separation}
\label{tab:comparison}
\centering
\renewcommand{\arraystretch}{1.18}
\setlength{\tabcolsep}{4.5pt}
\resizebox{\textwidth}{!}{%
\begin{tabular}{l  cccc  cccccccc}
\toprule
\multirow{2}{*}{\textbf{Model}}
  & \multicolumn{4}{c}{\textbf{Overall}}
  & \multicolumn{8}{c}{\textbf{By Reasoning Category}} \\
\cmidrule(lr){2-5} \cmidrule(l){6-13}
  & \textbf{Token F1}$\uparrow$& \textbf{ROUGE-L}$\uparrow$ & \textbf{METEOR}$\uparrow$ & \textbf{BERTScore-F1}$\uparrow$
  & \textbf{Path} & \textbf{Spat.} & \textbf{Attr.} & \textbf{Reg.}
  & \textbf{D.\,-O.} & \textbf{Occ.} & \textbf{H.\,-F.} & \textbf{V.Sep.} \\
\midrule
\multicolumn{13}{c}{\textit{Closed-source General VLMs}} \\[1pt]
GPT-5.4\cite{OpenAI2026GPT54SystemCard}
  & \textbf{38.15} & \textbf{39.58} & \textbf{37.44} & \underline{50.78} & \textbf{38.42} & \textbf{40.94} & \underline{38.25} & \textbf{33.81} & \textbf{43.14} & \textbf{29.68} & \underline{46.24} & \underline{41.25} \\
Gemini-3.1-Flash-Lite-Preview\cite{GoogleDeepMind2026Gemini31FlashLiteModelCard}
  & \underline{36.57} & \underline{35.67} & 30.45 & \textbf{55.89} & 34.71 & \underline{40.57} & \textbf{39.93} & \underline{32.87} & 38.52 & 25.75 & 42.33 & 38.98 \\
Claude-Opus-4.6\cite{Anthropic2026ClaudeOpus46}
  & 31.70 & 34.65 & \underline{30.78} & 45.11 & 27.32 & 35.49 & 22.29 & 28.42 & 33.41 & 23.03 & 36.93 & 33.71 \\
Perceptron-Mk1\cite{perceptronmk1_2026}
  & 34.24 & 30.96 & 26.99 & 44.05 & \underline{35.53} & 35.84 & 25.79 & 29.87 & \underline{39.74} & \underline{27.41} & \textbf{50.32} & \textbf{43.09} \\
Grok-4.3\cite{xAI2026Grok4}
  & 23.95 & 23.65 & 19.77 & 32.53 & 27.32 & 25.01 & 24.63 & 22.96 & 27.76 & 16.83 & 36.10 & 27.90 \\
\midrule
\multicolumn{13}{c}{\textit{Open-source General VLMs}} \\[1pt]
GLM-4.5V\cite{GLM45VTechnicalReport2025}
  & 27.86 & 27.04 & 20.85 & 38.12 & 30.90 & 27.85 & 30.53 & 25.81 & 34.81 & 21.23 & 42.75 & 40.73 \\
GLM-4.6V\cite{GLM45VTechnicalReport2025}
  & 29.75 & 30.57 & 25.44 & 42.11 & 30.00 & 31.19 & 30.78 & 29.31 & 36.32 & 18.39 & \underline{45.41} & \textbf{41.79} \\
LLaMA-3.2-90B-V\cite{MetaAI2024LLaMA32ModelCard}
  & \underline{33.60} & \underline{34.69} & 31.22 & 40.99 & 30.59 & \underline{37.19} & 32.71 & \underline{31.73} & 33.43 & 24.66 & 35.78 & 34.51 \\
Qwen3-VL-8B-Instruct\cite{Qwen3-VL}
  & 32.83 & 34.22 & \underline{36.46} & 46.02 & \underline{33.05} & 34.90 & \underline{35.44} & 28.62 & \underline{37.63} & \underline{24.68} & 44.58 & \underline{41.22} \\
MiMo-v2.5\cite{mimov25}
  & 25.89 & 26.79 & 29.22 & 23.20 & 24.06 & 26.66 & 18.24 & 25.97 & 23.04 & 23.78 & 30.28 & 32.19 \\
Nemotron-Nano-Omni-30B-A3B-R.\cite{nvidia2026nemotron3nanoomni}
  & 17.88 & 20.57 & 16.32 & 28.45 & 17.77 & 19.50 & 18.53 & 17.79 & 19.30 & 11.04 & 20.65 & 22.76 \\
Kimi-K2.5\cite{KimiTeam2026KimiK2.5}
  & 24.33 & 25.75 & 21.38 & 39.77 & 23.39 & 25.44 & 20.17 & 21.89 & 30.74 & 17.30 & 41.23 & 31.67 \\
Step-3.7-Flash\cite{StepFun2025Step3.7Flash}
  & 16.14 & 18.18 & 14.06 & 24.98 & 16.37 & 15.51 & 20.54 & 16.31 & 18.82 & 12.09 & 33.12 & 25.81 \\
MiniMax-01\cite{minimax2025minimax01scalingfoundationmodels}
  & \textbf{39.53} & \textbf{40.62} & \textbf{39.86} & \textbf{50.15} & \textbf{37.47} & \textbf{42.51} & \textbf{39.13} & \textbf{39.22} & \textbf{42.44} & \textbf{28.23} & \textbf{52.98} & 40.34 \\
ERNIE-4.5-VL-424B-A47B\cite{ernie45_2025}
  &  23.50 & 24.39 & 20.76 & 41.00 & 23.16 & 23.92 & 16.79& 22.64 & 29.95 & 19.28 & 36.59 & 23.57  \\
Nex-N2-Pro\cite{nexn2pro_2026}
  & 26.08 & 28.57 & 25.03 & 40.82 & 26.96 & 27.60 & 24.47 & 24.40 & 28.83 & 20.06 & 34.78 & 28.92 \\
Seed-2.0-Lite\cite{seed2lite_2026}
  & 30.77 & 33.35 & 30.42 & \underline{46.53} & 30.96 & 34.53 & 24.28 & 26.55 & 30.99 & 20.70 & 44.33 & 35.53\\
\midrule
\multicolumn{13}{c}{\textit{Spatial-reasoning VLMs}} \\[1pt]
SpaceLLaVA-1.5-7b\cite{chen2024spatialvlm}
  & 49.66 & 50.01 & 47.65 & 56.66 & 45.54 & 51.23 & 55.33 & 48.89 & 45.63 & 30.33 & \underline{60.69} & \underline{59.66} \\
SpatialRGPT-VILA1.5-8B\cite{cheng2024spatialrgpt}
  & \underline{53.75} & \underline{55.2} & \underline{50.47} & \underline{60.22} & \underline{58.99} & \underline{65.41} & \underline{56.47} & \underline{52.63} & \underline{56.11} & \underline{35.14} & 49.54 & 55.69 \\
\textbf{SatAgent (Ours)}
  & \textbf{65.44} & \textbf{66.10} & \textbf{62.01} & \textbf{72.31} & \textbf{65.00} & \textbf{72.45} & \textbf{62.98} & \textbf{58.90}
  & \textbf{70.44} & \textbf{43.54} & \textbf{78.51} & \textbf{71.70} \\
\bottomrule
\end{tabular}}
\end{table*}

\subsection{Implementation Details}
\label{sec:impl}

For the parameter settings of our SatAgent method, all input images
are uniformly resized and normalized, and the maximum text sequence
length is set to 192 tokens. The visual encoder is CLIP ViT-B/16
\cite{Radford2021CLIP}, and the depth estimation backbone is UniDepth
\cite{piccinelli2024unidepth}, both of which serve with their
parameters frozen during training, with only the projection layer
$\mathrm{clip\_proj}$ left trainable. The language backbone is Qwen3.5-9B,
loaded in 4-bit double quantization with NF4 format; LoRA adapters
are applied to its attention projection layers
($\mathbf{W}_Q,\mathbf{W}_K,\mathbf{W}_V,\mathbf{W}_O$) and
feed-forward network layers (gate, up, and down projections), with
rank $r{=}16$, scaling factor $\alpha_{\text{LoRA}}{=}32$, and dropout
rate $0.1$, accounting for approximately $1\%$ of the total
parameters. During training, labels for question tokens are masked to
$-100$, ensuring that the language modeling loss is computed only for
answer tokens.

During training, we train the model for 15 epochs with an effective
batch size of 16 (per-device batch size of 4 with a gradient
accumulation step of 4), using the fused AdamW optimizer
(\texttt{adamw\_torch\_fused}). The learning rate follows a cosine
annealing schedule with a 200-step linear warm-up and an initial
learning rate of $1\times10^{-4}$. Training employs FP16 mixed
precision with TF32 acceleration enabled for matrix operations, and a
gradient clipping threshold of $1.0$. In the loss function, the
auxiliary loss weight coefficients are set to $\lambda_1{=}0.1$,
$\lambda_2{=}0.2$, and $\lambda_s{=}0.05$, tuned via the validation
loss curve; evaluation is performed every 500 steps, and the
checkpoint with the lowest validation loss is retained as the final
model. At inference, the maximum number of newly generated tokens is
set to 64, and structured prompts constrain the output format
(yes/no or single-word/phrase) according to question category,
eliminating the need for constrained decoding. All experiments are
conducted on two NVIDIA 4090 GPUs.

\subsection{Comparisons with SOTA on SatAgent-SR130K}
To comprehensively evaluate SatAgent, we benchmark it against three groups of baselines on the SatAgent-SR130K test set: (i) closed-source general VLMs (GPT-5.4, Gemini-3.1-Flash-Lite-Preview, Claude-Opus-4.6, Perceptron-Mk1, Grok-4.3); (ii) open-source general VLMs (GLM-4.5V/4.6V, LLaMA-3.2-90B-V, Qwen3-VL-8B-Instruct, MiniMax-01, ERNIE-4.5-VL-424B-A47B, Seed-2.0-Lite, Nex-N2-Pro, etc.); and (iii) dedicated spatial-reasoning VLMs (SpaceLLaVA, SpatialRGPT). All models are evaluated zero-shot under an identical prompting protocol and inference pipeline using Token F1, ROUGE-L, METEOR, and BERTScore-F1.

As shown in Table~\ref{tab:comparison}, SatAgent achieves an overall Token~F1 of 65.44\%, surpassing the previous state-of-the-art dedicated spatial reasoning model SpatialRGPT by 11.69\% and the strongest general-purpose VLM MiniMax-01 by 25.91\%. This gain originates from the functional differentiation of the Dual-Channel Collaborative Encoder: by explicitly separating ventral semantic processing from dorsal geometric encoding, the model simultaneously achieves global semantic understanding and local geometric perception---a combination unavailable to any existing single-perspective method.
 
General-purpose VLMs exhibit a characteristic semantic-strong, geometry-weak performance profile. Models such as Qwen3.5-9B and GPT-5.4 perform reasonably on categories supported by appearance-based priors (e.g., Height-Footprint Consistency) but degrade substantially on geometry-intensive tasks such as Occlusion Completion, where SatAgent leads GPT-5.4 by 13.86\%. This gap directly exposes the absence of a dorsal-pathway equivalent in existing VLMs---a deficit that SatAgent's geometry-aware 3D reconstruction encoder bridges by lifting oblique UAV features into depth-coordinated BEV representations via 3D Gaussian soft projection.

To adapt these baselines to our cross-view evaluation protocol, SpaceLLaVA was evaluated with the drone oblique image as the sole visual input and satellite BEV context supplied via a structured system prompt. SpatialRGPT was deployed in region-free mode, replacing explicit bounding-box proposals with a full-image region token. Dedicated spatial reasoning models narrow this gap through spatial priors, yet both remain single-perspective architectures incapable of establishing cross-view topological correspondences such as path connectivity and building adjacency. SatAgent's Multi-view Topology-Semantic Alignment Module addresses this directly, surpassing SpatialRGPT by 7.04\% on Spatial Relation and 6.01\% on Path Reachability. The advantage is most pronounced on cross-view reasoning tasks (55.96\% vs.\ 46.22\% for SpatialRGPT), where coordinated ventral--dorsal feedback is essential for maintaining physically consistent alignment under extreme viewpoint disparities.

\begin{table*}[t]
\centering
\caption{Fine-Tuned VLM Baselines on SatAgent-SR130K: Accuracy (Token F1, ROUGE-L, METEOR, BERTScore-F1) and inference cost (latency, throughput, peak VRAM) of zero-shot and LoRA fine-tuned (FT) Qwen3.5 backbones (0.8B, 2B, 4B, 9B) versus SatAgent on SatAgent-SR130K.\textbf{Bold}: best;
\underline{Underline}: second best.}
\label{tab:finetune_baselines}
\renewcommand{\arraystretch}{1.3}   % 增大行距
\begin{tabular}{l cccc ccc}
\toprule
\multirow{2}{*}{\textbf{Model}} & \multicolumn{4}{c}{\textbf{Accuracy}} & \multicolumn{3}{c}{\textbf{Inference Cost}} \\
\cmidrule(lr){2-5} \cmidrule(lr){6-8}
 & Token F1 $\uparrow$ & ROUGE-L $\uparrow$ & METEOR $\uparrow$ & BERTScore-F1 $\uparrow$ & 
Latency$_{p50}$ (ms) $\downarrow$ & Throughput (tok/s) $\uparrow$ & Peak VRAM (MB) $\downarrow$ \\
\midrule
Qwen3.5-0.8B        & 22.77 & 23.37 & 19.87 & 28.49 & 384.9 & 20.3 & 1719 \\
Qwen3.5-0.8B-FT     & 48.82 & 50.56 & 50.05 & 59.63 & 682.1 & 21.7 & 1901 \\
Qwen3.5-2B          & 31.23 & 32.82 & 29.74 & 44.68 & 498.4 & 24.1 & 4302 \\
Qwen3.5-2B-FT       & 48.86 & 50.76 & 50.00 & 60.09 & 701.1 & 20.1 & 4682 \\
Qwen3.5-4B          & 33.65 & 36.18 & 34.93 & 44.97 & 934.8 & 26.4 & 8808 \\
Qwen3.5-4B-FT       & 49.74 & 51.08 & 49.94 & 60.86 & 949.7 & 15.9 & 9488 \\
Qwen3.5-9B          & 34.12 & 35.65 & 32.49 & 47.05 & 1357 & 15.9 & 17721 \\
Qwen3.5-9B-FT       & \underline{50.51} & \underline{52.43} & \underline{50.42} & \underline{61.91} & 1464.5 & 15.6 & 18833 \\
\midrule
\textbf{SatAgent (ours)} &   \textbf{65.44} & \textbf{66.10} & \textbf{62.01} & \textbf{72.31} & \textbf{1596} & \textbf{13.5} & \textbf{19056}\\
\bottomrule
\end{tabular}
\end{table*}

\begin{table}[t]
\centering
\caption{Parameter Counts of Fine-Tuned VLM and SatAgent}
\label{tab:params}
\renewcommand{\arraystretch}{1.3}
\begin{tabular}{l ccc}
\toprule
\textbf{Model} & \textbf{Total (M)} & \textbf{Trainable (M)} & \textbf{Trainable \%} \\
\midrule
Qwen3.5-0.8B-FT & 859.38 & 12.78 & 1.5 \\
Qwen3.5-2B-FT   & 2224.15 & 21.82 & 0.98 \\
Qwen3.5-4B-FT   & 4560.5 & 42.46 & 0.93 \\
Qwen3.5-9B-FT   & 9438.9 & 58.20 & 0.62 \\
\midrule
\textbf{SatAgent (Ours)} & \textbf{9446} & \textbf{64.25} & \textbf{0.68} \\
\bottomrule
\end{tabular}
\end{table}
\subsection{Fine-Tuned VLM Baselines on SatAgent-SR130K}
We compare the performance and inference cost of zero-shot and LoRA fine-tuned (FT) Qwen3.5 models (0.8B--9B) with SatAgent on the SatAgent-SR130K test set(Table~\ref{tab:finetune_baselines}). Across all scales, fine-tuning improves Token F1 by up to 26.05\%, BERTScore-F1 by up to 31.14\%, indicating that domain-specific supervision substantially narrows the gap between general-purpose vision-language models and cross-view spatial reasoning. Meanwhile, simply scaling up model size already yields diminishing marginal returns: although Qwen3.5-9B-FT has roughly 11$\times$ as many parameters as Qwen3.5-0.8B-FT, its F1 is still 14.93 \% lower than that of SatAgent. We also report the inference cost of the VLM baselines and SatAgent: compared with Qwen3.5-9B, the most expensive backbone, SatAgent's p50 latency increases by around 130 ms, throughput decreases by 2 tokens/s, and peak VRAM usage changes by around 200 MB, indicating that the additional overhead introduced by SatAgent relative to the backbone models is negligible. SatAgent achieves the best overall accuracy at an inference cost comparable to the mid-sized backbones, demonstrating that the primary driver of the performance gains is our architectural design, rather than fine-tuning or parameter scale alone.
Table~\ref{tab:params} compares total and trainable parameters of the LoRA-tuned Qwen3.5 backbones (0.8B--9B) and SatAgent. All models retain trainable ratios below 1.5\%. SatAgent's total parameters (9446M) align with its Qwen3.5-9B backbone, and its trainable budget (64.25M) exceeds Qwen3.5-9B-FT (58.20M) by merely 6.05M---the architecture-specific modules beyond LoRA---indicating its performance gains stem primarily from architectural design rather than added capacity.
% ============================================================
%  Comparison & Ablation Tables --- SatAgent
%  Dependencies: booktabs, multirow, makecell, array
% ============================================================

% ============================================================
% D. Ablation Study
% ============================================================
 
\subsection{Ablation Study}
 
To systematically validate the effectiveness of SatAgent's design,
we conduct hyperparameter analysis and three groups of progressive ablation experiments on the SatAgent-SR130K test set---analyzing the necessity of multi-view input, the key modules of geometry-aware encoding and structural alignment,
and the contribution of dual-pathway functional separation, respectively.
All variants are trained independently under identical hyperparameters and training epochs,
with the Overall Token F1 (\%) as the primary evaluation metric.
Category abbreviations follow the conventions of Table~\ref{tab:comparison}.

\begin{figure}[t!]
    \centering
    \includegraphics[width=1\linewidth]{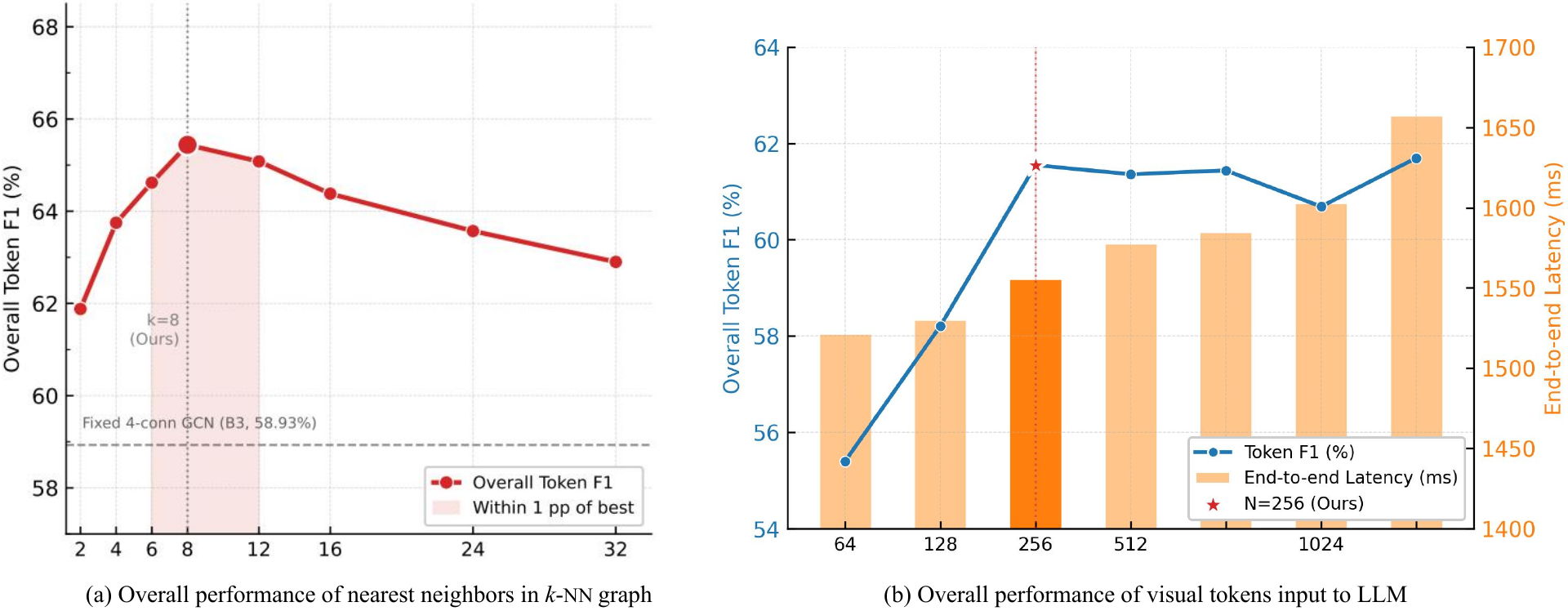}
    \caption{Hyperparameter sensitivity analysis.
    (a) Effect of the number of k-NN neighbors on overall Token F1;
    (b) Effect of the number of visual tokens input to the LLM on Token F1.}
    \label{fig:Hyperparameter}
\end{figure}
\noindent\textbf{Group~I: Selection of Key Hyperparameters}

We examine two hyperparameters that differ fundamentally in
how they affect the model, and are therefore evaluated under distinct
protocols, as shown in Fig.~\ref{fig:Hyperparameter}. The $k$-NN neighbor count $k$ in the semantically adaptive
topology layer only reshapes the sparse attention mask without
changing the matrix-multiplication cost, primarily affecting the
model's topological modeling capacity (too small a $k$ may omit
important neighbors, while too large a $k$ may introduce noisy
connections). We therefore warm-start from the final SatAgent
checkpoint; overall Token F1 peaks at $k{=}8$ and remains within
$1\%$ of the optimal value over $k\in[6,12]$, while $k<4$ degrades performance sharply and $k>16$ approaches the fixed four-connected
GCN baseline. The visual token count $N_v$ is a \textit{structural}
parameter, so its value must be fixed \textit{before} training the
final model used in the SOTA comparison. We screen $N_v \in [64, 2048]$
via shortened full retraining on a training subset, and observe that
$N_v \geq 512$ yields only marginal further gains (${<}1\%$). We
ultimately adopt $N_v{=}256$, which achieves the best trade-off
between accuracy and inference latency.

% ------------------------------------------------------------------
% GROUP I
% ------------------------------------------------------------------
\begin{table}[t!]
  \renewcommand{\arraystretch}{1.2}
  \caption{Group~II: Necessity of Multi-View Input.\textbf{Avg.}: mean over the four categories shown.}
  \label{tab:ablation_group2}
  \centering
  \footnotesize
  \setlength{\tabcolsep}{4pt}
  \begin{tabular}{l l c c c c c}
    \toprule
    ID & Variant & \textbf{Avg.} & Path & Spat. & H.\text{-}F. & V.Sep. \\
    \midrule
    A1 & Drone Only           & 43.15 & 38.50 & 45.21 & 46.44 & 45.90 \\
    A2 & Concat-2View         & 48.22 & 45.77 & 50.23 & 48.55 & 47.50 \\
    \midrule
    \textbf{Ours} & \textbf{SatAgent} & \textbf{71.92} & \textbf{65.00} & \textbf{72.45} & \textbf{78.51} & \textbf{71.70} \\
    \bottomrule
  \end{tabular}
\end{table}
\noindent\textbf{Group~II: Necessity of Multi-View Collaborative Input}
 
This group validates the following core questions:
whether UAV-Satellite dual-perspective collaboration is a necessary condition for reliable spatial reasoning,
and whether the geometry-aware fusion scheme proposed in this paper offers substantial gains over naive feature concatenation.
To this end, we retain two reference variants: \textbf{A1 (Drone Only)} zeros out the satellite branch features,
leaving the model to rely solely on the UAV local perspective for reasoning, lacking global topological priors;
\textbf{A2 (Concat-2View)} directly concatenates the mean-pooled CLIP global embeddings of the satellite and UAV images
for input to the LLM without any BEV projection or coordinate system calibration,
serving as a fair comparison baseline to evaluate the gains of the proposed geometry-aware fusion scheme.

% ---- Group II results ----
The results are shown in Table~\ref{tab:ablation_group2}.
The most pronounced degradation of A1 occurs in the Path Reachability category (declining by 26.50\% relative to SatAgent),
since path connectivity judgment requires a global topological view across regions,
and the limited field of view of UAVs---constrained by flight altitude and pitch angle---cannot reliably infer from a single oblique perspective whether a path is connected across occluded regions;
this result directly corroborates the irreplaceable role of the satellite top-view perspective as a global topological anchor.
A2 improves over A1 by 5.07\% overall, indicating that the introduction of the satellite perspective does carry effective information;
however, the gap between A2 and the full model remains most pronounced in Height-Footprint Consistency (declining by 29.96\%)
and Vertical Separation (declining by 24.20\%),
which require three-dimensional spatial structure perception,
indicating that directly concatenating satellite and UAV features at the semantic level fails to elicit depth and height information---explicit geometric modeling in the BEV coordinate system is a necessary prerequisite for effectively exploiting the 3D cues from the UAV perspective,
and cannot be replaced by mere semantic feature superimposition.
 
% ------------------------------------------------------------------
% GROUP II
% ------------------------------------------------------------------
\begin{table*}[!t]
  \renewcommand{\arraystretch}{1.2}
  \caption{Group~III: Key Design Components of Geometry-Aware BEV Encoding and Semantic Topology Alignment.
    Parentheses describe the core difference between each variant and the complete model.\textbf{Avg.}: mean over the four categories shown.}
  \label{tab:ablation_group3}
  \centering
  \small
  \begin{tabular}{llccccc}
    \toprule
    ID & Variant & \textbf{Avg.} & Path & Occ. & H.\text{-}F. & V.Sep. \\
    \midrule
    B1 & w/o BEVScaleCalibrator  & 54.41 & 50.44 & 37.56 & 65.88 & 63.77 \\
    B2 & w/o UniDepth   & 55.53 & 60.45 & 35.66 & 63.56 & 62.43 \\
    B3 & Fixed 4-conn.\ GCN      & 58.93 & 61.56 & 32.76 & 75.50 & 65.88 \\
    \midrule
    \textbf{Ours} & \textbf{SatAgent} & \textbf{64.69} & \textbf{65.00} & \textbf{43.54} & \textbf{78.51} & \textbf{71.70} \\
    \bottomrule
  \end{tabular}
\end{table*}
 
\begin{table*}[!t]
  \renewcommand{\arraystretch}{1.3}
  \caption{Group~IV: Ablation Results for Dual-Pathway Functional Separation.
    Task accuracy is measured by \textbf{Token F1 (\%)}.\textbf{Avg.}: mean over the five categories shown.
    $\mathcal{L}_{\text{spec}}$ refers to the specialization loss functions $\mathcal{L}_{\text{geo}} + \mathcal{L}_{\text{div}}$.}
  \label{tab:ablation_group4}
  \centering
  \small
  \begin{tabular}{cl c cccc c}
    \toprule
    ID & Variant & \textbf{Avg.} & Spat. & Attr. & D.\text{-}O. & Occ. & H.\text{-}F. \\
    \midrule
    C1 & Baseline (Single CLIP Stream) & 56.89 & 55.49 & 62.10 & 61.55 & 35.20 & 70.12  \\
    C2 & w/o Ventral Adapter (V-Stream) & 60.37 & 57.66 & 65.42 & 66.12 & 38.45 & 74.20 \\
    C3 & w/o Dorsal Adapter (D-Stream) & 60.16 & 59.32 & 68.85 & 63.90 & 36.90 & 71.85 \\
    C4 & w/o Bidirectional Gating & 64.43  & 62.05 & 71.20 & 69.34 & 42.12 & 77.45  \\
    C5 & w/o Spec. Losses ($\mathcal{L}_{\text{spec}}$) & 58.60 & 56.25 & 64.88 & 63.20 & 36.75 & 71.90  \\
    \midrule
    \textbf{Ours} & \textbf{SatAgent} & \textbf{63.98} & \textbf{72.45} & \textbf{62.98} & \textbf{70.44} & \textbf{43.54} & \textbf{78.51} \\
    \bottomrule
  \end{tabular}
\end{table*}
\noindent\textbf{Group~III: Key Design of Geometry-Aware BEV Encoding and Semantic Topology Alignment}

This group isolates three core components of the geometry-aware encoder and the semantically adaptive topology alignment module. \textbf{B1 (w/o BEVScaleCalibrator)} removes the affine coordinate calibration module, fusing UAV BEV features with satellite features without scale/offset compensation, simulating uncorrected systematic bias between the two coordinate systems. \textbf{B2 (w/o UniDepth)} replaces the UniDepth depth backend with a lightweight CNN of comparable parameter count, isolating the impact of depth quality on BEV reconstruction. \textbf{B3 (Fixed 4-conn.\ GCN)} replaces the dynamic $k$-NN graph and multi-head attention with a standard fixed four-connected GCN, degrading the topology layer to isotropic low-pass filtering.

As shown in Table~\ref{tab:ablation_group3}, each variant exhibits category-differentiated degradation that maps directly onto its module's function. B1 degrades most on Path Reachability ($-14.56\%$), since path-connectivity judgment depends on precise spatial correspondence between UAV BEV and satellite features, and coordinate misalignment introduces systematic displacement that disrupts cross-view topological association; Occlusion Completion, which relies more on local depth than on global alignment, declines only $-5.98\%$, confirming that calibration benefit scales with a task's spatial-precision demand. B2 degrades most on Height-Footprint Consistency ($-14.95\%$) and Occlusion Completion ($-7.88\%$), indicating that high-quality monocular depth is a prerequisite for effective 3D Gaussian reconstruction: degraded depth maps yield geometrically meaningless covariance parameters, causing the soft projection to collapse toward noise-like aggregation. B3 degrades most on Occlusion Completion ($-10.78\%$), which requires multi-region associative reasoning, while Height-Footprint Consistency drops only $-3.01\%$---the fixed four-connected grid aggregates only spatially adjacent nodes and cannot link semantically related but spatially non-contiguous regions (e.g., same-type buildings in different lots), quantitatively validating the advantage of the dynamic $k$-NN graph for semantic topological connectivity.

\begin{figure*}[!t]
    \centering
    \includegraphics[width=0.75\linewidth]{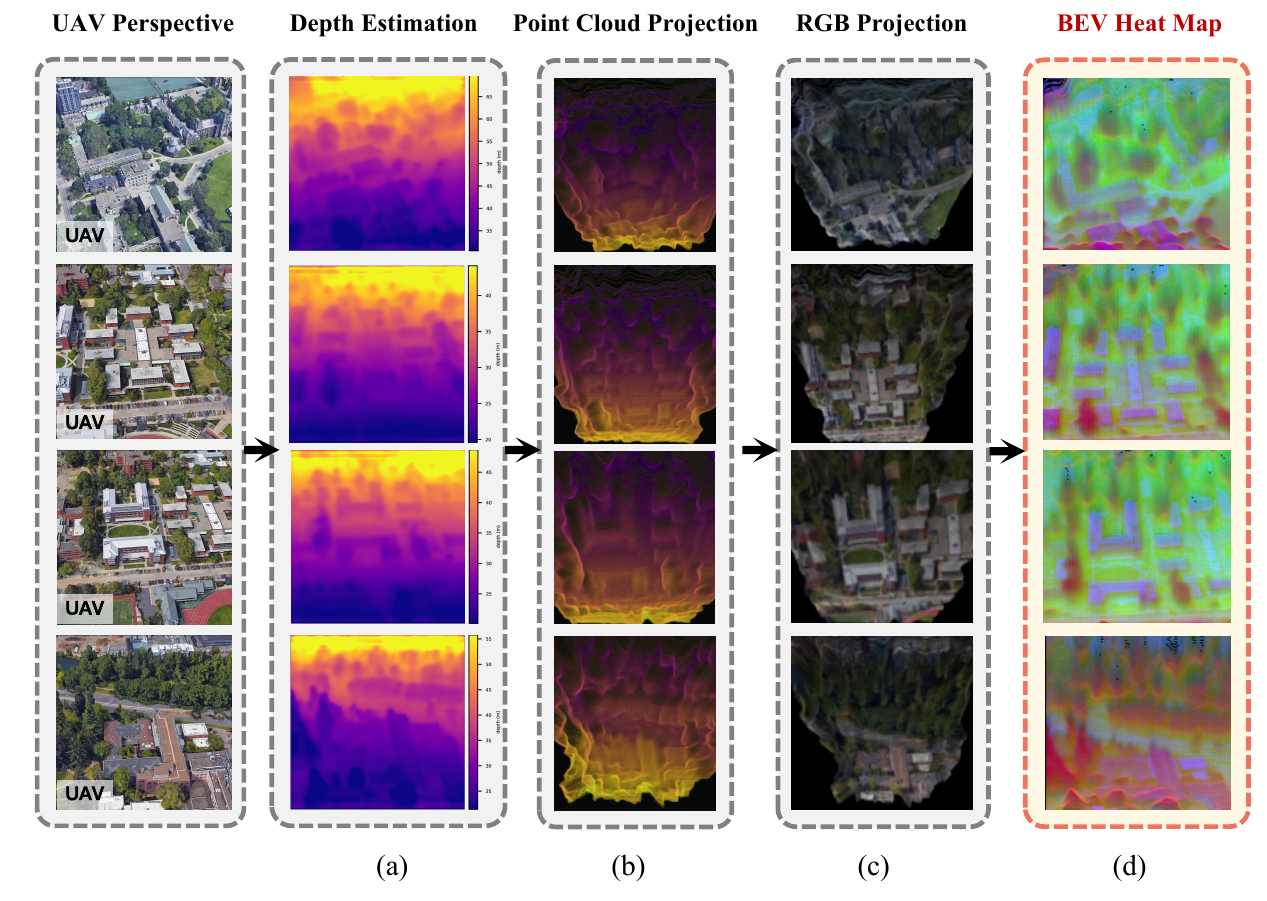}
    \caption{Complete encoding pipeline visualization of the geometry-aware 3D reconstruction encoder:
(a) per-pixel depth estimation by UniDepth;
(b) 3D point cloud back-projection based on learnable camera intrinsic parameters;
(c) RGB texture BEV map generated by EWA soft projection;
(d) three-channel BEV feature map encoding point density (R), relative height corrected via orthographic rectification and ground-plane fitting (G), and mean brightness (B), providing a joint geometric-appearance representation for cross-view feature alignment.}
    \label{fig:visualize}
\end{figure*}
 
\begin{figure*}[!t]
    \centering
    \includegraphics[width=0.85\linewidth]{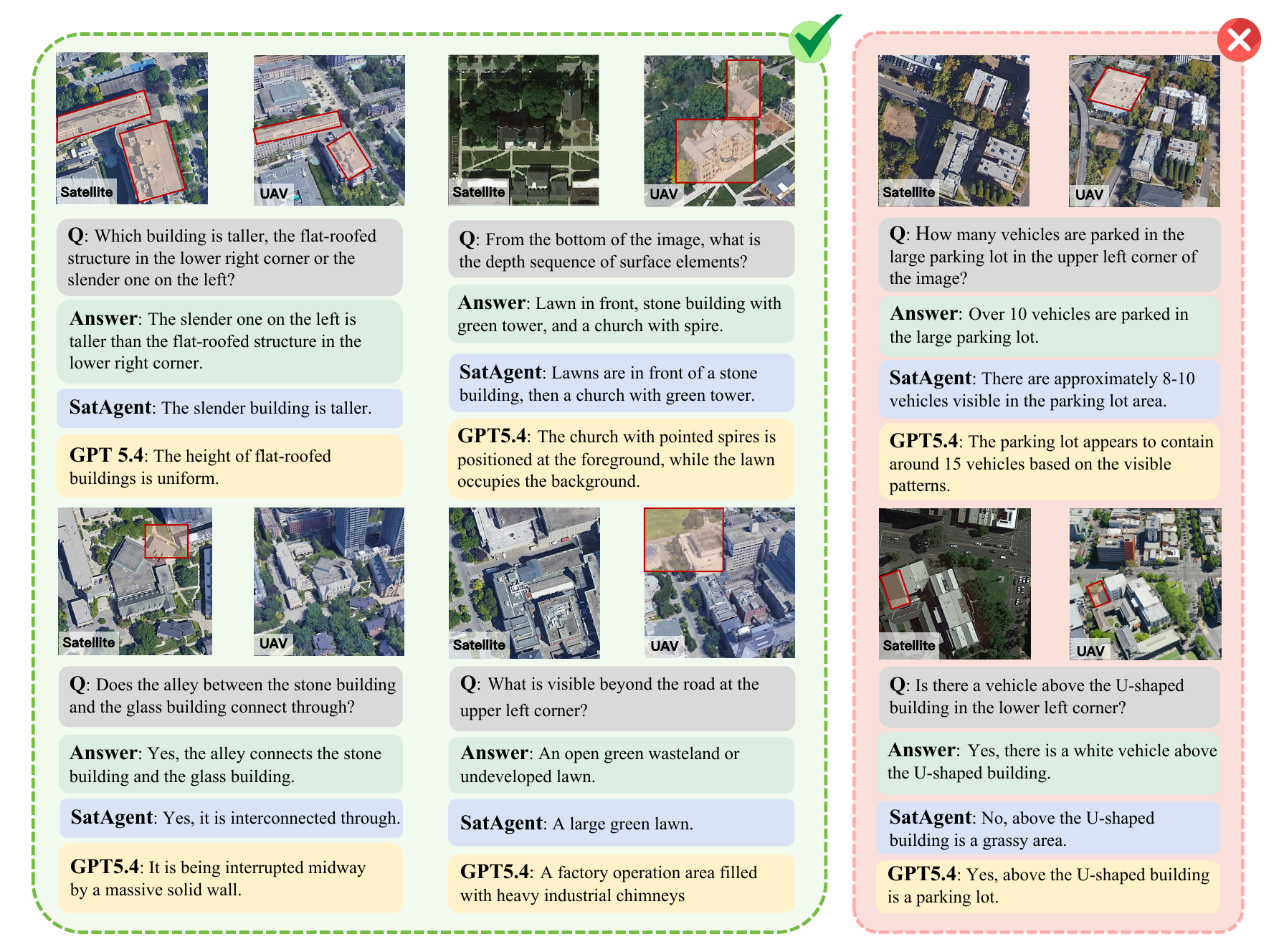}
    \caption{Qualitative comparison of SatAgent and GPT-5.4 on 3D spatial reasoning tasks; the third column shows failure cases.}
    \label{fig:compare_gpt}
\end{figure*}
% ------------------------------------------------------------------
% GROUP III
% ------------------------------------------------------------------
\noindent\textbf{Group~IV: Contribution of Dual-Pathway Functional Separation}

This group validates the dual-pathway functional separation design from both structural and training-signal perspectives, using Token F1 as the metric. The five variants: \textbf{C1 (Baseline, Single CLIP Stream)} removes all dual-pathway components, with both branches degraded to shared CLIP encoding; \textbf{C2 (w/o Ventral Adapter)} removes the ventral adapter, so the satellite branch degrades to standard CLIP projection without semantic prototype attention; \textbf{C3 (w/o Dorsal Adapter)} removes the dorsal adapter, so the UAV branch degrades to standard CLIP projection without geometric feature injection; \textbf{C4 (w/o Bidirectional Gating)} retains both adapters but removes cross-stream gating, processing the branches independently; \textbf{C5 (w/o Spec.\ Losses)} retains the full dual-pathway structure but removes $\mathcal{L}_{\text{geo}}+\mathcal{L}_{\text{div}}$, isolating structural from training-signal contributions.

Table~\ref{tab:ablation_group4} reveals four findings. \textit{(i) The dual-pathway architecture is necessary}: C1 lags SatAgent by $7.49\%$--$10.35\%$ across all categories, while C2/C3---retaining a single adapter---recover to within $3.61\%$--$3.82\%$, showing the structure itself contributes independently of the specialization losses. \textit{(ii) Category-specific degradation confirms the ventral$\to$semantics, dorsal$\to$geometry assignment}: despite similar overall drops, C2 degrades disproportionately on semantic tasks (Attr.\ $-7.03\%$, Spat.\ $-5.32\%$) while geometric tasks stay relatively intact, whereas C3 shows the inverse (H.-F.\ $-6.66\%$, Occ.\ $-6.64\%$, D.-O.\ $-6.54\%$ vs.\ Attr.\ $-3.60\%$, Spat.\ $-3.66\%$), providing direct evidence of pathway specialization. \textit{(iii) Bidirectional gating yields consistent but modest gains}: C4 trails SatAgent uniformly by $0.93\%$--$1.42\%$, consistent with its near-zero gating-strength initialization. \textit{(iv) Specialization losses are indispensable}: C5, which keeps the dual-pathway structure but removes $\mathcal{L}_{\text{geo}}$ and $\mathcal{L}_{\text{div}}$, drops $6.61\%$--$7.57\%$ per category---substantially more than C2 or C3 alone---confirming that explicit gradient pressure from these losses is required to realize the full benefit of dual-pathway differentiation.

% ============================================================
%  Qualitative Analysis --- SatAgent
% ============================================================

% ============================================================
%  4.5  Qualitative Analysis
% ============================================================
\subsection{Qualitative Analysis}

Fig.~\ref{fig:visualize} presents intermediate representation
visualizations of the geometry-aware 3D reconstruction encoder across
four representative scenes. Starting from UAV RGB images, the encoder
estimates per-pixel absolute metric depth maps via UniDepth
(column~a), where clear depth discontinuities at building edges
provide a reliable geometric foundation for 3D reconstruction. Pixels
are then back-projected into 3D point clouds (column~b) and projected
onto a horizontal BEV grid with RGB texture (column~c), visually
approximating satellite orthographic imagery and validating the
oblique-to-top-down geometric transformation. The final three-channel
BEV feature map (column~d) encodes point density~(R), relative
height~(G), and mean brightness~(B); perspective trapezoid distortion
is eliminated via orthographic correction
$X_{\text{ortho}}{=}X{\cdot}(Z_{\text{ref}}/Z)$, and relative height
is extracted through linear ground-plane fitting
$H{=}Z{-}Z_{\text{ground}}(Y)$, causing building rooftops to be
prominently distinguished from the ground in the G channel.

Fig.~\ref{fig:compare_gpt} presents a qualitative comparison between
SatAgent and GPT-5.4 across four representative spatial reasoning
tasks: height--footprint consistency, depth-order consistency, path
reachability, and distant region identification. In all four cases,
GPT-5.4 produces geometrically implausible answers---reversing
foreground-background ordering, hallucinating obstructions, or
misidentifying land-cover categories at range---whereas SatAgent
yields correct responses by grounding language generation on the
explicit depth, height, and density channels encoded in the BEV
representation. These results demonstrate that the geometry-aware 3D
reconstruction encoder confers a systematic advantage on tasks where
metric spatial structure is decisive.

% ============================================================
%  4.6  Failure Cases and Discussion
% ============================================================
\subsection{Failure Cases and Discussion}

Despite its strong overall performance, SatAgent exhibits two
systematic failure modes under extreme conditions, illustrated in
column~three of Fig.~\ref{fig:compare_gpt}.

The visual token compression required to control LLM sequence length
causes spatial over-smoothing in high-density scenes, degrading
fine-grained instance localization such as precise vehicle counting.
When multiple instances occupy spatially adjacent BEV cells, the
$4{\times}4$ adaptive pooling merges their representations into a
single token, losing the inter-instance discriminability needed for
exact enumeration.

Large temporal gaps between satellite and UAV acquisitions introduce
irreconcilable semantic conflicts during cross-view alignment: surface
changes such as construction or vegetation growth produce feature
distributions in one view that contradict the structural priors
established by the other, causing cross-view gating to produce
unstable fusion weights and ultimately degrading reasoning accuracy in
dynamically changed areas.

Both failure modes share a common underlying tension between
computational efficiency and spatial fidelity. Token compression
trades representational resolution for LLM tractability; temporal
misalignment trades data acquisition flexibility for cross-view
consistency. Notably, these failure modes differ qualitatively from
those of single-perspective baselines such as SpatialRGPT, which
degrade due to absent geometric grounding rather than resolution or
temporal limitations---suggesting that SatAgent's failures arise at a
higher level of the reasoning hierarchy where multi-view collaboration
is already established but fidelity constraints become binding.
We plan to address these limitations in future work through adaptive
token allocation conditioned on local feature entropy, and
temporal-aware alignment that explicitly models surface change between
acquisition timestamps.

% ============================================================
%  5  Conclusion
% ============================================================
\section{Conclusion}

This paper proposes SatAgent, a UAV-Satellite collaborative spatial
reasoning model that addresses two fundamental limitations of
existing single-perspective VLMs: severe perspective distortion and
the absence of explicit geometric modeling. By incorporating satellite
orthographic imagery as a global structural reference alongside UAV
oblique imagery, and constructing a physically grounded feature fusion
pipeline---comprising dual-pathway collaborative encoding,
covariance-aware 3D Gaussian BEV reconstruction, and multi-view
topology-semantic alignment---SatAgent eliminates cross-view semantic
misalignment within a unified metric coordinate system, endowing the
LLM with spatial perception capabilities grounded in the real physical
world. Evaluated on the purpose-built SatAgent-SR130K benchmark,
SatAgent comprehensively surpasses existing general-purpose VLMs and
dedicated spatial reasoning models across all eight reasoning
categories. We plan to extend the model in future work
to an
``air--space--ground'' tri-perspective architecture incorporating
ground-level street views, explore adaptive token allocation and
temporal-aware alignment to address the identified failure modes, and
investigate large-scale pre-training to further improve
generalization.

\bibliographystyle{IEEEtran}
\bibliography{main}

% ----------------------------- Author Biographies -----------------------------
% ----------------------------- Author Biographies -----------------------------
{\setlength{\parskip}{4pt}\setlength{\parindent}{0pt}

\textbf{Wenyi Zhang} received the B.Sc. degree from Chongqing University, Chongqing, China, in 2024. He is currently a Ph.D. student with the Aerospace Information Research Institute, Chinese Academy of Sciences, Beijing, China.

\textbf{Fanglong Yao} received the Ph.D. degree from the Aerospace Information Research Institute, Chinese Academy of Sciences, Beijing, China, in 2022. He is currently an Associate Professor with the Aerospace Information Research Institute, Chinese Academy of Sciences, Beijing, China.

\textbf{Youzhi Liu} received the B.Sc. degree from Hunan University, Changsha, China, in 2022. He is currently a Ph.D. student with the Aerospace Information Research Institute, Chinese Academy of Sciences, Beijing, China.

\textbf{Peng Hu} received the B.Sc. degree from Beihang University, Beijing, China, in 2019. He is currently a Ph.D. student with the School of Computer Science and Engineering, Beihang University, Beijing, China.

\textbf{Zhengqiu Zhu} received the Ph.D. degree in management science and engineering from the National University of Defense Technology, Changsha, China, in 2023. He is currently an Associate Professor with the College of Systems Engineering, National University of Defense Technology, Changsha, China.

\textbf{Chen Gao} received the B.Sc. and Ph.D. degrees from the Department of Electronic Engineering, Tsinghua University, Beijing, China. He is currently a Faculty Member (Research-track AP) of BNRist, Tsinghua University, Beijing, China.

\textbf{Xian Sun} received the B.Sc. degree from Beihang University (Beijing University of Aeronautics and Astronautics), Beijing, China, in 2004, and the M.Sc. and Ph.D. degrees from the Institute of Electronics, Chinese Academy of Sciences, Beijing, China, in 2009. He is currently a Professor with the Aerospace Information Research Institute, Chinese Academy of Sciences, Beijing, China.

\textbf{Kun Fu} received the B.Sc., M.Sc., and Ph.D. degrees from the National University of Defense Technology, Changsha, China. He is currently a Professor with the Aerospace Information Research Institute, Chinese Academy of Sciences, Beijing, China.

}

\clearpage
\appendices
\section{Baseline Model Descriptions}

\noindent\textbf{GPT-5.4}~\cite{OpenAI2026GPT54SystemCard}: OpenAI's flagship closed-source multimodal model, used as a general-purpose reasoning baseline with strong world knowledge but limited explicit geometric grounding.

\noindent\textbf{Gemini-3.1-Flash-Lite-Preview}~\cite{GoogleDeepMind2026Gemini31FlashLiteModelCard}: Google DeepMind's lightweight multimodal model, balancing efficiency and reasoning as an efficient closed-source baseline.

\noindent\textbf{Claude-Opus-4.6}~\cite{Anthropic2026ClaudeOpus46}: Anthropic's high-capability multimodal model, representative of closed-source general VLMs with strong language understanding but weak metric spatial reasoning.

\noindent\textbf{Perceptron-Mk1}~\cite{perceptronmk1_2026}: Perceptron AI's vision-language model specialized for video and embodied spatial reasoning, included as a closed-source spatially-aware baseline.

\noindent\textbf{Grok-4.3}~\cite{xAI2026Grok4}: xAI's general-purpose multimodal model, included as an additional closed-source baseline for broad comparison on spatial tasks.

\noindent\textbf{GLM-4.5V / GLM-4.6V}~\cite{GLM45VTechnicalReport2025}: Zhipu AI's open-source vision-language model series, representing mainstream Chinese open-weight multimodal baselines with general visual understanding.

\noindent\textbf{LLaMA-3.2-90B-V}~\cite{MetaAI2024LLaMA32ModelCard}: Meta's large open-source multimodal model, providing a strong general-purpose open-weight baseline for visual reasoning.

\noindent\textbf{Qwen3-VL-8B-Instruct}~\cite{Qwen3-VL}: Alibaba's instruction-tuned open-source vision-language model, representing a compact, efficient open-weight multimodal baseline.

\noindent\textbf{MiMo-v2.5}~\cite{mimov25}: Xiaomi's open-source multimodal model, included as an additional general-purpose open-weight baseline.

\noindent\textbf{Nemotron-Nano-Omni-30B-A3B-R.}~\cite{nvidia2026nemotron3nanoomni}: NVIDIA's compact omni-modal MoE model, evaluated as an efficient open-source baseline with reasoning-oriented post-training.

\noindent\textbf{Kimi-K2.5}~\cite{KimiTeam2026KimiK2.5}: Moonshot AI's open-source multimodal model, providing a general-purpose open-weight baseline for visual-language reasoning.

\noindent\textbf{Step-3.7-Flash}~\cite{StepFun2025Step3.7Flash}: StepFun's lightweight open-source multimodal model, included as an efficiency-oriented open-weight baseline.

\noindent\textbf{MiniMax-01}~\cite{minimax2025minimax01scalingfoundationmodels}: MiniMax's open-source multimodal model with long-context support, serving as a strong general-purpose open-weight baseline.

\noindent\textbf{ERNIE-4.5-VL-424B-A47B}~\cite{ernie45_2025}: Baidu's large-scale MoE multimodal model, representing a heavyweight open-source baseline with heterogeneous text--vision expert routing.

\noindent\textbf{Nex-N2-Pro}~\cite{nexn2pro_2026}: Nex AGI's MoE model post-trained from Qwen3.5 for agentic tasks, included as a reasoning-oriented open-source baseline.

\noindent\textbf{Seed-2.0-Lite}~\cite{seed2lite_2026}: ByteDance's omni-modal understanding model, evaluated as an efficient open-source baseline with unified video/image/text capability.

\noindent\textbf{SpaceLLaVA-1.5-7b}~\cite{chen2024spatialvlm}: A spatial-reasoning-tuned VLM built on LLaVA, incorporating spatial VQA data to improve 3D relation understanding from single images.

\noindent\textbf{SpatialRGPT-VILA1.5-8B}~\cite{cheng2024spatialrgpt}: A region-aware spatial reasoning VLM that injects depth and bounding-box region tokens for fine-grained metric spatial inference.

\section{Inter-Category Correlation Analysis}

\begin{figure}[htbp]
  \centering
  \includegraphics[width=0.9\linewidth]{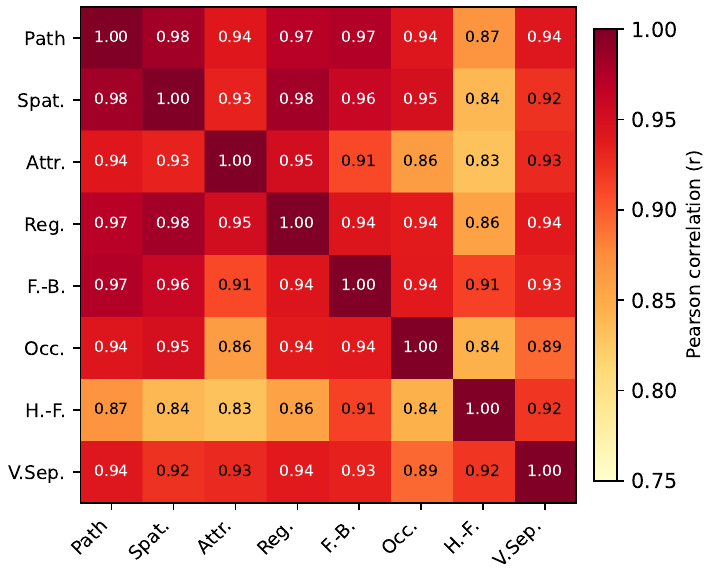}
  \caption{Pearson correlation matrix of Token F1 scores across the eight reasoning categories in Table~\ref{tab:comparison}, computed over 20 evaluated models (19 baselines + SatAgent).}
  \label{fig:correlation_matrix}
\end{figure}

To examine whether the eight reasoning categories in Table~\ref{tab:comparison} probe distinct or overlapping capabilities, we compute the $8\times8$ Pearson correlation matrix of Token~F1 scores across all 20 evaluated models (19 baselines plus SatAgent), treating each model's per-category score as one observation. Fig.~\ref{fig:correlation_matrix} visualizes the resulting matrix.

Overall, correlations are uniformly high ($r=0.83$--$0.98$), indicating that a dominant common factor---general cross-view reasoning capability---governs performance across all categories: models that excel on one task tend to excel on most others, and vice versa. Path Reachability, Spatial Relation, and Region Identification form the most tightly coupled cluster ($r\geq0.97$), consistent with their shared reliance on global topological correspondence between the two views. Height--Footprint Consistency (H.-F.) is comparatively the most distinct category, showing the lowest correlations with Attribute Grounding ($r=0.83$), Spatial Relation ($r=0.84$), and Occlusion Completion ($r=0.84$); this category can be partially solved via appearance-based priors (e.g., building height cues from shadows or façade texture) even by models lacking explicit geometric reasoning, making it a comparatively weaker proxy for overall spatial-reasoning ability. These results support treating the eight categories as a coherent but non-redundant benchmark suite.

\section{Confidence Intervals for Table~\ref{tab:comparison}}
\label{app:ci}

Table~\ref{tab:ci_overall} reports the key metrics results of Table~\ref{tab:comparison} with 95\% bootstrap confidence-interval half-widths (1{,}000 resamples over the test set) appended as $\pm$ values.

\begin{table*}[p]
\centering
\vspace*{\fill}
\caption{95\% bootstrap CI half-widths (1000 resamples) for the four overall metrics in Table~\ref{tab:comparison}.}
\label{tab:ci_overall}
\footnotesize
\renewcommand{\arraystretch}{1.15}
\setlength{\tabcolsep}{6pt}
\begin{tabular}{l cccc}
\toprule
\textbf{Model} & \textbf{Token F1} & \textbf{ROUGE-L} & \textbf{METEOR} & \textbf{BERTScore-F1} \\
\midrule
\multicolumn{5}{c}{\textit{Closed-source General VLMs}} \\[1pt]
GPT-5.4                        & $38.15\pm1.35$   & $39.58\pm1.24$   & $37.44\pm1.31$   & $50.78\pm1.28$ \\
Gemini-3.1-Flash-Lite-Preview  & $36.57\pm1.54$   & $35.67\pm1.61$   & $30.45\pm1.55$   & $55.89\pm1.66$ \\
Claude-Opus-4.6                & $31.70\pm1.22$   & $34.65\pm1.14$   & $30.78\pm1.03$   & $45.11\pm1.25$ \\
Perceptron-Mk1                 & $34.24\pm1.66$ & $30.96\pm1.56$ & $26.99\pm1.47$ & $44.05\pm1.28$ \\
Grok-4.3                       & $23.95\pm1.17$ & $23.65\pm1.07$ & $19.77\pm0.98$ & $32.53\pm1.19$ \\
\midrule
\multicolumn{5}{c}{\textit{Open-source General VLMs}} \\[1pt]
GLM-4.5V                       & $27.86\pm1.44$   & $27.04\pm1.39$   & $20.85\pm1.47$   & $38.12\pm1.50$ \\
GLM-4.6V                       & $29.75\pm1.56$ & $30.57\pm1.47$ & $25.44\pm1.40$ & $42.11\pm1.50$ \\
LLaMA-3.2-90B-V                & $33.60\pm1.82$   & $34.69\pm1.74$   & $31.22\pm1.52$   & $40.99\pm1.33$ \\
Qwen3-VL-8B-Instruct           & $32.83\pm1.47$ & $34.22\pm1.44$ & $36.46\pm1.65$ & $46.02\pm1.29$ \\
MiMo-v2.5                      & $25.89\pm1.37$ & $26.79\pm1.41$ & $29.22\pm1.63$ & $23.20\pm1.94$ \\
Nemotron-Nano-Omni-30B-A3B-R.  & $17.88\pm1.36$ & $20.57\pm1.34$ & $16.32\pm1.31$ & $28.45\pm1.51$ \\
Kimi-K2.5                      & $24.33\pm1.42$ & $25.75\pm1.34$ & $21.38\pm1.21$ & $39.77\pm1.39$ \\
Step-3.7-Flash                 & $16.14\pm1.39$ & $18.18\pm1.45$ & $14.06\pm1.31$ & $24.98\pm1.93$ \\
MiniMax-01                     & $39.53\pm1.48$ & $40.62\pm1.49$ & $39.86\pm1.60$ & $50.15\pm1.21$ \\
ERNIE-4.5-VL-424B-A47B         & $23.50\pm1.46$ & $24.39\pm1.39$ & $20.76\pm1.35$ & $41.00\pm1.25$ \\
Nex-N2-Pro                     & $26.08\pm1.31$ & $28.57\pm1.14$ & $25.03\pm1.17$ & $40.82\pm1.17$ \\
Seed-2.0-Lite                  & $30.77\pm1.43$ & $33.35\pm1.34$ & $30.42\pm1.36$ & $46.53\pm1.16$ \\
\midrule
\multicolumn{5}{c}{\textit{Spatial-reasoning VLMs}} \\[1pt]
SpaceLLaVA-1.5-7b               & $49.66\pm1.26$   & $50.01\pm1.69$   & $47.65\pm1.55$   & $56.66\pm1.47$ \\
SpatialRGPT-VILA1.5-8B           & $53.75\pm1.41$   & $55.20\pm1.27$   & $50.47\pm1.53$   & $60.22\pm1.89$ \\
\midrule
\textbf{SatAgent (Ours)}        & $65.44\pm1.62$   & $66.10\pm1.52$   & $62.01\pm1.74$   & $72.31\pm1.42$ \\
\bottomrule
\end{tabular}
\vspace*{\fill}
\end{table*}


\end{document}